\documentclass[10pt,twocolumn,letterpaper]{article}

\usepackage[pagenumbers]{cvpr} %

\usepackage[dvipsnames,table]{xcolor}

\newcommand{\method}{AEROBLADE}

\usepackage{multirow}
\usepackage{subcaption}
\usepackage{stfloats}
\usepackage{siunitx}
\usepackage[accsupp]{axessibility} %

\let\cline\cmidrule

\definecolor{cvprblue}{rgb}{0.21,0.49,0.74}
\usepackage[pagebackref,breaklinks,colorlinks,citecolor=cvprblue]{hyperref}

\def\paperID{00000} %
\def\confName{CVPR}
\def\confYear{2024}

\title{AEROBLADE: Training-Free Detection of Latent Diffusion Images Using Autoencoder Reconstruction Error}

\author{Jonas Ricker \qquad Denis Lukovnikov \qquad Asja Fischer\\
Ruhr University Bochum\\
{\tt\small \{jonas.ricker, denis.lukovnikov, asja.fischer\}@rub.de}
}

\begin{document}

\maketitle

\begin{abstract}
With recent text-to-image models, anyone can generate deceptively realistic images with arbitrary contents, fueling the growing threat of visual disinformation. A key enabler for generating high-resolution images with low computational cost has been the development of latent diffusion models (LDMs). In contrast to conventional diffusion models, LDMs perform the denoising process in the low-dimensional latent space of a pre-trained autoencoder (AE) instead of the high-dimensional image space. Despite their relevance, the forensic analysis of LDMs is still in its infancy. In this work we propose AEROBLADE, a novel detection method which exploits an inherent component of LDMs: the AE used to transform images between image and latent space. We find that generated images can be more accurately reconstructed by the AE than real images, allowing for a simple detection approach based on the reconstruction error. Most importantly, our method is easy to implement and does not require any training, yet nearly matches the performance of detectors that rely on extensive training. We empirically demonstrate that AEROBLADE is effective against state-of-the-art LDMs, including Stable Diffusion and Midjourney. Beyond detection, our approach allows for the qualitative analysis of images, which can be leveraged for identifying inpainted regions. We release our code and data at \href{https://github.com/jonasricker/aeroblade}{https://github.com/jonasricker/aeroblade}.
\end{abstract}

\begin{figure*}[ht]
\centering
    \begin{subfigure}{0.15\textwidth}
        \centering
        \includegraphics{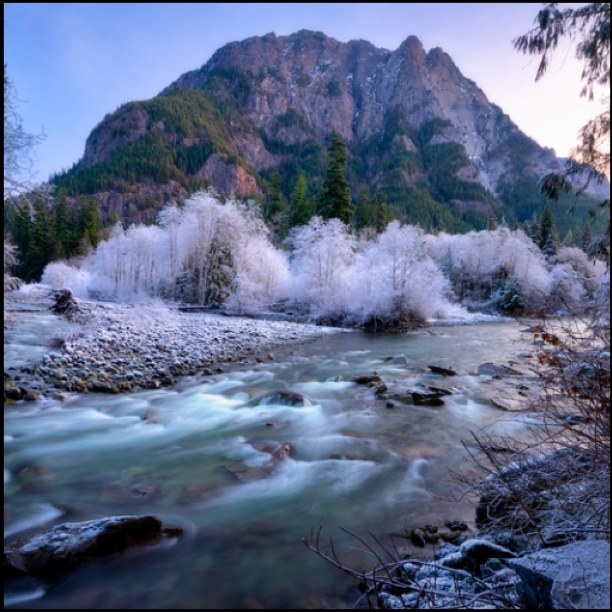}
        \caption{$x_\text{real}$}
    \end{subfigure}
    \begin{subfigure}{0.15\textwidth}
        \centering
        \includegraphics{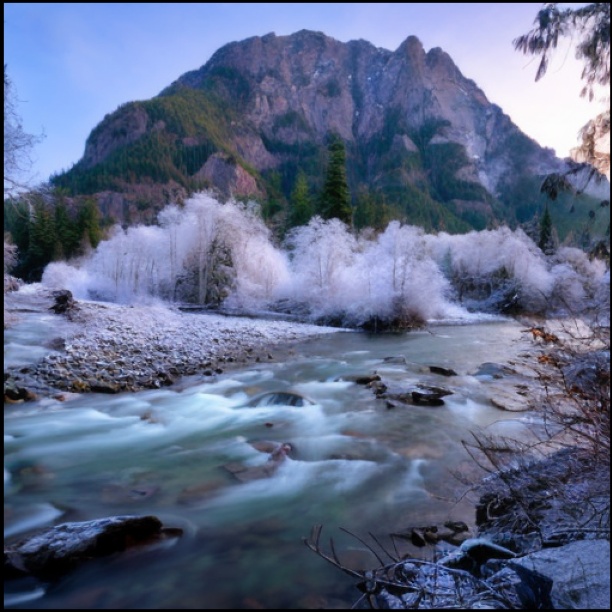}
        \caption{$\mathcal{D}(\mathcal{E}(x_\text{real}))$}
    \end{subfigure}
    \begin{subfigure}{0.15\textwidth}
        \centering
        \includegraphics{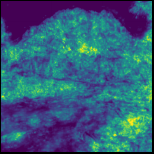}
        \caption{$\Delta_{\text{AE}_i}(x_\text{real})$}
        \label{fig:lpips_example:real_error}
    \end{subfigure}
    \begin{subfigure}{0.15\textwidth}
        \centering
        \includegraphics{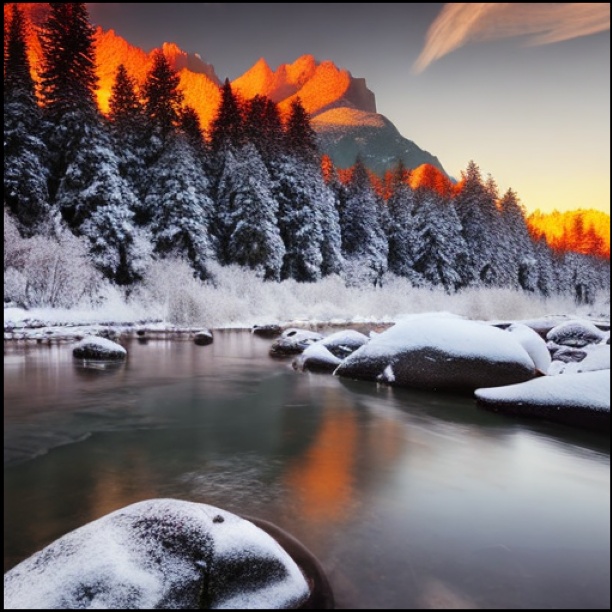}
        \caption{$x_\text{SD2.1}$}
    \end{subfigure}
    \begin{subfigure}{0.15\textwidth}
        \centering
        \includegraphics{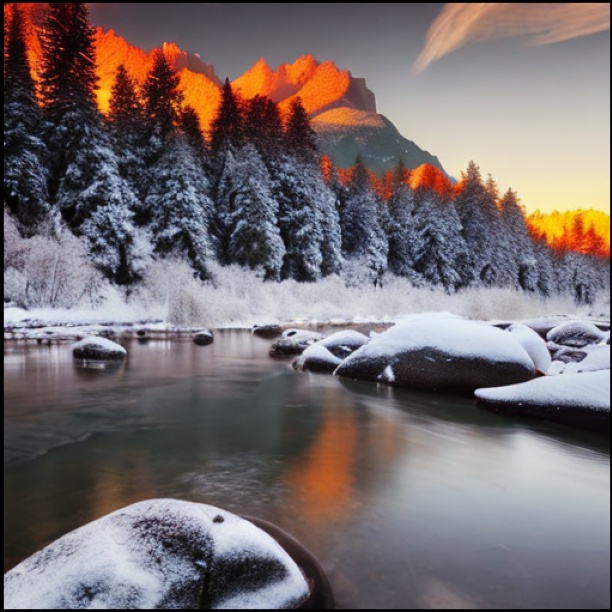}
        \caption{$\mathcal{D}(\mathcal{E}(x_\text{SD2.1}))$}
    \end{subfigure}
    \begin{subfigure}{0.15\textwidth}
        \centering
        \includegraphics[trim=0 0 25 0, clip]{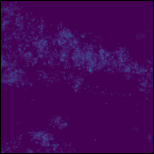}
        \caption{$\Delta_{\text{AE}_i}(x_\text{SD2.1})$}
    \end{subfigure}
    \begin{subfigure}{0.05\textwidth}
        \centering
        \includegraphics[trim=76 0 0 0, clip]{assets/figures/lpips_example/fake_lpips.pdf}
        \caption*{}
    \end{subfigure}
    \caption{Example illustrating the idea behind \method{}. (a) shows a real image from LAION-5B~\citep{schuhmannLAION5BOpenLargescale2022} and (d) is generated by Stable Diffusion 2.1~\citep{rombachHighresolutionImageSynthesis2022}. (b) and (e) are the corresponding reconstructions obtained by passing the original images through the AE of Stable Diffusion 2. (c) and (f) visualize the error between original and reconstruction measured using the LPIPS~\citep{zhangUnreasonableEffectivenessDeep2018} distance. The reconstruction error is significantly lower for the generated image $x_\text{SD2.1}$ than for the real image $x_\text{real}$, which \method{} leverages for detection.}
    \label{fig:lpips_example}
\end{figure*}

\section{Introduction}\label{sec:intro}
The emergence of powerful and easy-to-use text-to-image models, like Stable Diffusion~\citep{stablediffusion} and Midjourney~\citep{midjourney}, marked a turning point in the history of generative AI. While generating high-quality images previously required technological expertise and significant computational resources, hyperrealistic scenes with arbitrary contents are now only a few keystrokes away. These models open up countless creative possibilities, but may also have harmful consequences, like generated pictures winning art contests~\citep{rooseGeneratedPictureWon2022} or fake photos of the Pope going viral~\citep{vincentSwaggedoutPopeAI2023}. While the ability to generate realistic images from prompts carries the inherent danger of malicious acts, the greater risk lies in the growing \textit{erosion of trust} in legitimate sources due to the flood of synthetic media~\citep{volpicelliAIEndPhotographic2023}. Developing effective and efficient detection methods is therefore of utmost importance.

Scaling the generation process to high resolutions while keeping the computational cost low was mainly made possible by so-called latent diffusion models (LDMs)~\citep{rombachHighresolutionImageSynthesis2022}. Despite their ubiquity, LDMs have not yet been sufficiently studied from the perspective of forensics. Whereas standard diffusion models (DMs) operate directly in the image space, LDMs use the low-resolution latent space of a pre-trained autoencoder (AE).
The model first generates a small latent representation, which is then transformed to a high-resolution image by the AE's decoder.

In this work, we demonstrate that this property allows for a simple yet highly effective approach for detecting images generated by LDMs. Our proposed method, \method{} (\textbf{a}uto\textbf{e}ncoder \textbf{r}ec\textbf{o}nstruction-\textbf{b}ased \textbf{la}tent diffusion \textbf{de}tection), distinguishes real and generated images by computing their AE reconstruction error, which is the distance between the original image and its reconstruction obtained by passing it through the AE's encoder and decoder (see \cref{sec:method}). While generated images can be accurately reconstructed, real images exhibit deficiencies, especially in complex regions (see \cref{fig:lpips_example}). By computing the reconstruction error for the AEs of multiple LDMs, our method is effective against a wide range of models and can be easily extended to new ones. We find that this simple approach is able to reliably detect generated images, achieving a mean \textbf{average precision (AP) of $\mathbf{0.992}$} on various state-of-the-art models including Stable Diffusion~\citep{stablediffusion}, Kandinsky~\citep{razzhigaevKandinskyImprovedTexttoimage2023}, and Midjourney~\citep{midjourney} (see \cref{sec:experiments:detection}). Notably, our method \textbf{does not require any training}, yet performs almost as well as extensively trained classifiers (see \cref{sec:experiments:comparison}). In contrast to most existing detectors, which usually output only a score denoting how likely it is an image is generated, \method{} additionally provides rich qualitative information, giving insights into how well certain regions can be reconstructed. We show that this information can be utilized for localizing inpainted regions within real images (see \cref{sec:experiments:qualitative}).

In summary, we make the following contributions:
\begin{itemize}
    \item We present \method{}, a simple and training-free method for detecting LDM-generated images based on the AE reconstruction error.
    \item We empirically show that our approach can effectively distinguish real images from images generated by seven state-of-the-art LDMs.
    \item Moreover, we thoroughly study the properties of AE reconstruction errors and demonstrate the qualitative insights \method{} provides, which can help to identify inpainted regions.
\end{itemize}

\section{Related Work}\label{sec:related_work}
\paragraph{Detection of Generated Images}
In consequence of the rapid evolution of generative AI, different approaches for detecting synthetic images have been explored in the recent past. One class of detection methods exploits visual artifacts, like incorrect illumination~\citep{maternExploitingVisualArtifacts2019,faridLightingConsistencyPaint2022}, inconsistent eye reflections~\citep{huExposingGANGeneratedFaces2021}, or irregular pupil shapes~\citep{guoEyesTellAll2022}. Another successful strategy is to analyze images in the frequency domain, where generated images exhibit distinguishable artifacts~\citep{frankLeveragingFrequencyAnalysis2020,chandrasegaranCloserLookFourier2021,schwarzFrequencyBiasGenerative2021,corviIntriguingPropertiesSynthetic2023}. Instead of using handcrafted features, a variety of learning-based methods has been proposed~\citep{rosslerFaceForensicsLearningDetect2019,cozzolinoUniversalGANImage2021,wangCNNgeneratedImagesAre2020,gragnanielloAreGANGenerated2021,ojhaUniversalFakeImage2023,corviDetectionSyntheticImages2023}. \citet{wangCNNgeneratedImagesAre2020} make the observation that a deep classifier trained to distinguish generated images by a single GAN from real images generalizes surprisingly well to images from other GANs. \citet{gragnanielloAreGANGenerated2021} show that using more extensive augmentation and omitting early downsampling improves the detection performance. Moreover, GAN inversion~\citep{xiaGANInversionSurvey2023} has been leveraged for identifying generated images~\citep{pasquiniIdentifyingSyntheticFaces2023}, which is related to the idea of reconstruction-based detection.

Most existing detectors are trained and evaluated on GAN-generated images, while the forensic analysis of DMs is still in an early stage~\citep{rickerDetectionDiffusionModel2023,shaDEFAKEDetectionAttribution2022,corviDetectionSyntheticImages2023,wangDIREDiffusiongeneratedImage2023,maexposing2023}. \citet{corviDetectionSyntheticImages2023} show that a deep classifier trained on images generated by the originally proposed LDM~\citep{rombachHighresolutionImageSynthesis2022} is able to detect images by this model, but has limited generalization capabilities. Aiming towards universal detection, \citet{ojhaUniversalFakeImage2023} propose to train a simple linear classifier on top of features from a pre-trained CLIP-ViT~\citep{dosovitskiyImageWorth16x162021,radfordLearningTransferableVisual2021}. They show that using a feature space not explicitly trained on real and fake images provides better generalization to new model classes, including DMs. Most related to the method proposed here is DIRE~\citep{wangDIREDiffusiongeneratedImage2023}. Similar to previous works, DIRE makes use of a deep network for classifying images as real or fake, however, it uses the differences between images and their reconstructions obtained from a pre-trained ADM~\citep{dhariwalDiffusionModelsBeat2021} as input data. Specifically, they use the deterministic forward and reverse process from DDIM~\citep{songDenoisingDiffusionImplicit2022} to map images to Gaussian noise and back. Our method differs considerably from DIRE, since \method{} does not require training a deep classifier and obtains the reconstruction error from just the AE, without going through the costly forward and backward processes. To the best of our knowledge, the only other training-free detection method for DMs is $\text{SeDID}_\text{Stat}$~\citep{maexposing2023}, which can be seen as a refined version of DIRE. Instead of using the difference between original and reconstruction, it exploits the difference between individual steps during the denoising process. Similar to DIRE and unlike our approach, $\text{SeDID}_\text{Stat}$ needs to perform computationally expensive diffusion and denoising operations. Moreover, it was only tested on low-resolution images ($32\times32$), and has shown to be sensitive to the choice of hyperparameters, in particular the step at which the error is computed.

\paragraph{Diffusion Models for Visual Anomaly Detection}
A different task for which the reconstruction capabilities of DMs can be used is visual anomaly detection. The idea is that anomalous regions of an image can be identified by reconstructing the image with a DM trained or conditioned on nominal images. Since the model can only generate data from the learned distribution of nominal samples, the difference between original and reconstruction reveals anomalies. This idea has been explored for general out-of-distribution detection~\citep{liuUnsupervisedOutofdistributionDetection2023}, medical images~\citep{wollebDiffusionModelsMedical2022}, industrial applications~\citep{mousakhanAnomalyDetectionConditioned2023}, and unsupervised video anomaly detection~\citep{turExploringDiffusionModels2023}.

\section{Preliminaries}\label{sec:preliminaries}
In this section we provide background information on LDMs and how they differ from standard DMs. We also explain the details of the LPIPS~\citep{zhangUnreasonableEffectivenessDeep2018} distance metric, which we use to estimate the reconstruction error.

\paragraph{Latent Diffusion Models (LDMs)}\label{sec:preliminaries:ldm}
DMs are a class of generative models based on nonequilibrium thermodynamics~\citep{sohl-dicksteinDeepUnsupervisedLearning2015} that have been shown to be capable of high-fidelity image synthesis~\citep{songGenerativeModelingEstimating2019,hoDenoisingDiffusionProbabilistic2020,dhariwalDiffusionModelsBeat2021,songDenoisingDiffusionImplicit2022,yangDiffusionModelsComprehensive2023}. In the forward (or diffusion) process, an image is gradually disturbed by adding Gaussian noise. The model, which typically uses the U-Net~\citep{ronnebergerUNetConvolutionalNetworks2015} architecture, then learns to recover a slightly less noisy version of the image at different steps. During the reverse (or denoising) process, new images are generated from pure Gaussian noise by reversing the diffusion process until a clean image is reached. Since predictions operate in the high-dimensional image space, both training and inference require excessive computational resources. To tackle this limitation, \citet{rombachHighresolutionImageSynthesis2022} propose to perform the denoising process in an optimized latent space of lower dimensionality. The key idea is to first generate the semantics of an image in latent space, which are then transformed to the image space, adding high-frequency details. Dividing the generation process into these two phases makes both training and sampling much more efficient. Several powerful text-to-image models are based on the concept of LDMs, including the popular Stable Diffusion~\citep{stablediffusion} and Midjourney~\citep{midjourney}.\footnote{If not stated otherwise, we use the term ``LDM'' to describe the class of DMs that perform denoising in latent space, not the particular LDM proposed by \citet{rombachHighresolutionImageSynthesis2022}.}

To map images from image to latent space and back, LDMs use a pre-trained AE. That is, during the forward process, an image $x$ is mapped to the latent representation $z=\mathcal{E}(x)$ using the encoder $\mathcal{E}$. During the reverse process, the final image $\tilde{x} = \mathcal{D}(\tilde{z})$ is obtained by passing the denoised latents $\tilde{z}$ through the decoder $\mathcal{D}$.

\paragraph{Learned Perceptual Image Patch Similarity (LPIPS)}\label{sec:preliminaries:lpips}
The LPIPS metric, proposed by \citet{zhangUnreasonableEffectivenessDeep2018}, aims to solve the problem of perceptual similarity, i.e., estimating how similar two images are according to human judgement. The authors make the observation that the internal activations of a classifier trained on ImageNet~\citep{russakovskyImageNetLargeScale2015} provide an embedding that corresponds surprisingly well with human perception, even across different architectures (SqueezeNet~\citep{iandolaSqueezeNetAlexNetlevelAccuracy2016}, AlexNet~\citep{krizhevskyImageNetClassificationDeep2012}, and VGG16~\citep{simonyanVeryDeepConvolutional2015}). They find that LPIPS outperforms specialized distance metrics like SSIM~\citep{wangImageQualityAssessment2004a} or MS-SSIM~\citep{wangMultiscaleStructuralSimilarity2003}.

To obtain the LPIPS distance between two images, both are fed through the network to extract the activations from certain layers. For VGG16, which we mainly use in this work, these are the five convolutional layers before the pooling layers. For each layer, the activations are scaled channel-wise using learned linear weights. The similarity is computed as the $\ell_2$ difference between both activations. The activations from each layer are then spatially averaged (along width and height), and the final similarity score is given as the sum of these averages. %

\section{Methodology}\label{sec:method}
We first introduce the general framework for reconstruction-based fake detection. Subsequently, we provide the details and formal definition of \method{}.

\paragraph{Reconstruction-based Fake Image Detection}\label{sec:methodology:framework}
The idea of reconstruction-based detection builds upon two assumptions. The first is that, given a generative model $G_i$ and an image $x$, we can compute a reconstruction $\tilde{x} = \phi_i(x)$, with $\phi_i$ denoting a reconstruction method that is based on $G_i$. The second assumption is that for an image $x_i$, which is generated by model $G_i$, the distance $d$ between the original image and its reconstruction, $d(x_i, \tilde{x}_i)$, is small. In contrast, a real image $x_r$ cannot be reconstructed as accurately, i.e., $d(x_r, \tilde{x}_r) > d(x_i, \tilde{x}_i)$.

\paragraph{\method{}}
The most crucial component of a reconstruction-based detection technique is the choice of the reconstruction method $\phi$.
We find that in the case of LDMs, there is a straightforward choice that is easy to implement and efficient. Our proposed method, \method{} (\textbf{a}uto\textbf{e}ncoder \textbf{r}ec\textbf{o}nstruction-\textbf{b}ased \textbf{la}tent diffusion \textbf{de}tection), is based on the observation that the model's AE is better at reconstructing generated images than reconstructing real images.
An intuitive explanation is that AEs use a latent representation $z$ that maps to a constrained data manifold. Generated images lie within it and can therefore be accurately reconstructed, while real images are mapped to the closest point in the manifold, leading to a higher error.
Therefore, the distance between an image and its reconstruction allows for simple threshold-based detection. In contrast to previous works~\citep{wangDIREDiffusiongeneratedImage2023,maexposing2023}, our method does neither require performing the costly deterministic denoising process, nor any additional training.

\begin{figure}[t]
    \centering
    \includegraphics[width=0.7\linewidth]{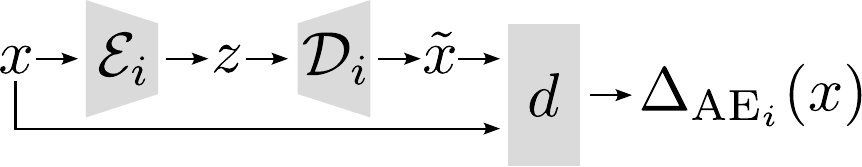}
    \caption{Graphical representation of \cref{eq:ae}. The reconstruction error $\Delta_{\text{AE}_i}(x)$ is defined as the distance between an image $x$ and its reconstruction $\tilde{x}$ obtained from passing it through the encoder $\mathcal{E}_i$ and decoder $\mathcal{D}_i$ of an LDM's AE.}
    \label{fig:aeroblade}
\end{figure}

We define the reconstruction error of an image $x$ based on the AE of an LDM $G_i$ (denoted by $\text{AE}_i$) as
\begin{equation}
    \Delta_{\text{AE}_i}(x) := d(x, \tilde{x}) = d(x, \mathcal{D}_i(\mathcal{E}_i(x)))\ ,
    \label{eq:ae}
\end{equation}
with $\mathcal{E}_i$ and $\mathcal{D}_i$ being the encoder and decoder of the AE, respectively, and $d$ being some distance metric.
We provide an illustration of \cref{eq:ae} in \cref{fig:aeroblade}.
In our experiments we find that LPIPS~\citep{zhangUnreasonableEffectivenessDeep2018} is a suitable choice for $d$, but we investigate alternative distance metrics in \cref{sec:experiments:additional}.

\cref{eq:ae} computes the reconstruction error using the AE from a single model, assuming that images are either real or generated by this particular model. In practice, however, with a constantly growing pool of generative models available, this assumption does not hold. To account for this, we compute $\Delta_{\text{AE}_i}$ for a set of LDMs, $\{G_i, i \in \mathcal{I}\}$, and use the smallest reconstruction error for classification.
Formally, we define the minimum reconstruction error as
\begin{equation}
    \Delta_\text{Min}(x) := \min_{i \in \mathcal{I}}\ \Delta_{\text{AE}_i} (x) = \min_{i \in \mathcal{I}}\ d(x, \mathcal{D}_i(\mathcal{E}_i(x)))\ .
    \label{eq:min}
\end{equation}
Based on the assumption that the ``correct'' AE provides the best reconstruction, using $\Delta_\text{Min}$ is more suitable for detection than, e.g., taking the average from multiple AEs.

\section{Experiments}\label{sec:experiments}
In this section we experimentally demonstrate the effectiveness of \method{}. After introducing the datasets and AEs we use in our evaluation, we report the detection performance and compare it to several state-of-the-art baselines. We finally demonstrate that our approach can be used to spot inpainted regions within real images and conduct further analyses regarding the properties of \method{}.

\subsection{Setup}
\paragraph{Data}
We evaluate our method on images from seven text-to-image LDMs, among them four open-source and three proprietary models. We include three different versions of Stable Diffusion~\citep{stablediffusion}, the initially released version 1.1 (SD1.1), 1.5 (SD1.5) which is widely used as a base for custom models by the generative AI community, and the more recent 2.1 (SD2.1). We additionally test on images generated by Kandinsky~2.1 (KD2.1)~\citep{razzhigaevKandinskyImprovedTexttoimage2023}, which builds upon Stable Diffusion and DALL-E~2~\citep{rameshHierarchicalTextconditionalImage2022}. To allow for a fair evaluation, we attempt to generate images whose contents match those of the real images (taken from LAION-5B~\citep{schuhmannLAION5BOpenLargescale2022}). To achieve this, we extract prompts from real images using CLIP interrogator~\citep{pharmapsychoticClipinterrogator2023}, which combines CLIP~\citep{radfordLearningTransferableVisual2021} and BLIP~\citep{liAlignFuseVision2021} to optimize a prompt towards a target image. Finally, we include data from three different versions of the popular image generation service Midjourney~\citep{midjourney}, namely versions 4 (MJ4), 5 (MJ5), and 5.1 (MJ5.1). Since these models are proprietary, we cannot extract matching prompts. Instead, we use a diverse set of images from the publicly available Midjourney Discord server.

We collect \num{1000} images for each generative model plus \num{1000} real images. We provide more details on the data collection in \cref{sup:data_collection} in the supplementary material.

\paragraph{Autoencoders}
We compute reconstructions using three different AEs: SD1 and SD2, which are used by Stable Diffusion models with the corresponding versions, and the AE from Kandinsky 2.1 (KD2.1). It should be noted that while all AEs have the same task, their architectures differ. Stable Diffusion uses a variational autoencoder (VAE)~\citep{kingmaAutoencodingVariationalBayes2022} with Kullback-Leibler regularization,
while Kandinsky uses a discrete vector quantized-VAE (VQ-VAE)~\citep{vandenoordNeuralDiscreteRepresentation2017}. We emphasize that the AE used by Midjourney is not publicly available, which allows us to test the applicability of our method to closed-source generators.

\subsection{Evaluation of Detection Performance}\label{sec:experiments:detection}
\begin{figure}
    \centering
    \includegraphics{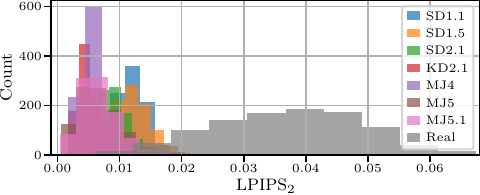}
    \caption{Distributions of reconstruction error $\Delta_\text{Min}$ using $\text{LPIPS}_2$ for different datasets. We provide results for all LPIPS variants in \cref{sup:histograms} in the supplementary material.}
    \label{fig:hist_conv2}
\end{figure}

We initially inspect the distributions of reconstruction errors for real and generated images. For each dataset, we compute the reconstruction errors from \num{1000} images using all three AEs and choose the minimum reconstruction error ($\Delta_\text{Min}$) per sample. Besides using the standard definition of LPIPS (summation of the Euclidean distances of the activations from all five layers), we also evaluate using the distances based on activations from a single layer only. We denote these as $\text{LPIPS}_i, i \in \{1,\dots,5\}$. The reconstruction errors for $\text{LPIPS}_2$ are shown in \cref{fig:hist_conv2}. We make the promising observation that generated images have a consistently lower reconstruction error than real images.

To obtain a quantitative evaluation of \method{}, we follow previous works~\citep{wangCNNgeneratedImagesAre2020, ojhaUniversalFakeImage2023, wangDIREDiffusiongeneratedImage2023} and report the performance in average precision (AP) in \cref{tab:lpips_layers}.
We first compare the results from different LPIPS variants.
Overall, our results suggest that the second LPIPS layer ($\text{LPIPS}_2$) captures the most meaningful differences between original images and reconstructions. The AP decreases towards higher layers (with a larger receptive field), indicating that fine-grained details cause higher reconstruction errors. We further analyze the relation between image complexity and reconstruction error in \cref{sec:experiments:qualitative}.

\begin{table}
    \setlength{\tabcolsep}{4.0pt}
    \centering
    \scriptsize
        \begin{tabular}{@{\ }llrrrrrrr@{\ }}
        \toprule
        Distance & AE & SD1.1 & SD1.5 & SD2.1 & KD2.1 & MJ4 & MJ5 & MJ5.1 \\
        \midrule
        \multirow[c]{4}{*}{LPIPS} & SD1 & 0.989 & 0.988 & 0.827 & 0.879 & 0.967 & 0.930 & 0.930 \\
         & SD2 & 0.763 & 0.771 & 0.992 & 0.878 & \textbf{0.999} & 0.994 & 0.994 \\
         & KD2.1 & 0.593 & 0.606 & 0.665 & 0.997 & 0.919 & 0.878 & 0.860 \\
         & Min & 0.959 & 0.957 & 0.991 & 0.996 & 0.998 & 0.993 & 0.994 \\
        \cline{1-9}
        \multirow[c]{4}{*}{$\text{LPIPS}_1$} & SD1 & 0.954 & 0.951 & 0.667 & 0.785 & 0.894 & 0.835 & 0.842 \\
         & SD2 & 0.574 & 0.593 & 0.967 & 0.790 & 0.995 & 0.979 & 0.980 \\
         & KD2.1 & 0.481 & 0.502 & 0.550 & 0.994 & 0.874 & 0.804 & 0.790 \\
         & Min & 0.882 & 0.884 & 0.961 & 0.994 & 0.993 & 0.975 & 0.976 \\
        \cline{1-9}
        \multirow[c]{4}{*}{$\text{LPIPS}_2$} & SD1 & \textbf{0.992} & \textbf{0.991} & 0.777 & 0.878 & 0.971 & 0.933 & 0.924 \\
         & SD2 & 0.716 & 0.722 & \textbf{0.996} & 0.879 & \textbf{0.999} & \textbf{0.998} & \textbf{0.997} \\
         & KD2.1 & 0.543 & 0.552 & 0.623 & \textbf{0.999} & 0.927 & 0.882 & 0.858 \\
         & Min & 0.979 & 0.978 & 0.994 & \textbf{0.999} & \textbf{0.999} & 0.997 & 0.996 \\
        \cline{1-9}
        \multirow[c]{4}{*}{$\text{LPIPS}_3$} & SD1 & 0.989 & 0.987 & 0.874 & 0.904 & 0.974 & 0.944 & 0.949 \\
         & SD2 & 0.828 & 0.832 & 0.991 & 0.905 & 0.997 & 0.993 & 0.995 \\
         & KD2.1 & 0.615 & 0.622 & 0.695 & 0.998 & 0.929 & 0.894 & 0.881 \\
         & Min & 0.969 & 0.965 & 0.991 & 0.997 & 0.997 & 0.993 & 0.995 \\
        \cline{1-9}
        \multirow[c]{4}{*}{$\text{LPIPS}_4$} & SD1 & 0.981 & 0.979 & 0.881 & 0.897 & 0.966 & 0.945 & 0.943 \\
         & SD2 & 0.841 & 0.845 & 0.983 & 0.896 & 0.997 & 0.989 & 0.991 \\
         & KD2.1 & 0.685 & 0.691 & 0.742 & 0.994 & 0.916 & 0.896 & 0.877 \\
         & Min & 0.926 & 0.924 & 0.983 & 0.990 & 0.996 & 0.989 & 0.991 \\
        \cline{1-9}
        \multirow[c]{4}{*}{$\text{LPIPS}_5$} & SD1 & 0.960 & 0.960 & 0.838 & 0.813 & 0.923 & 0.904 & 0.898 \\
         & SD2 & 0.800 & 0.814 & 0.965 & 0.815 & 0.989 & 0.976 & 0.978 \\
         & KD2.1 & 0.662 & 0.670 & 0.700 & 0.969 & 0.861 & 0.850 & 0.820 \\
         & Min & 0.867 & 0.869 & 0.964 & 0.944 & 0.989 & 0.976 & 0.978 \\
        \bottomrule
        \end{tabular}
    \caption{Detection performance of \method{} for different LPIPS variants (first column) and AEs (second column), measured in AP. ``Min'' refers to the minimum reconstruction error $\Delta_\text{Min}$, as defined in~\cref{eq:min}. The remaining columns contain the results for individual datasets. The best result for each dataset is highlighted in \textbf{bold}.}
    \label{tab:lpips_layers}
\end{table}

\begin{table}
    \setlength{\tabcolsep}{4.0pt}
    \centering
    \scriptsize
    \begin{tabular}{@{\ }llrrrrrrr@{\ }}
        \toprule
        Distance & AE & SD1.1 & SD1.5 & SD2.1 & KD2.1 & MJ4 & MJ5 & MJ5.1 \\
        \midrule
        \multirow[c]{3}{*}{$\text{LPIPS}_2$} & SD1 & \textbf{1.000} & \textbf{1.000} & 0.000 & 0.000 & 0.000 & 0.000 & 0.000 \\
         & SD2 & 0.000 & 0.000 & \textbf{1.000} & 0.000 & \textbf{0.999} & \textbf{0.997} & \textbf{0.995} \\
         & KD2.1 & 0.000 & 0.000 & 0.000 & \textbf{1.000} & 0.001 & 0.003 & 0.005 \\
        \bottomrule
        \end{tabular}
    \caption{Fraction of samples for which an AE has the smallest reconstruction error using $\text{LPIPS}_2$. The highest fraction for each dataset is highlighted in \textbf{bold}. We provide results for all LPIPS variants in \cref{sup:attribution} in the supplementary material.}
    \label{tab:attribution_conv2}
\end{table}

Second, we compare the performance of different AEs to obtain the reconstructions.
We observe that, as was to be expected, using the ``matching'' AE for a given dataset (e.g., SD1 for SD1.1) yields the best results, with the AP ranging from $0.991$ to $0.999$ (with $\text{LPIPS}_2$). However, as the results for Midjourney demonstrate, accurate detection is possible even without access to the generator's AE. While the AP is relatively high for all three AEs, it appears that the AE used by Midjourney is similar to the one used by Stable Diffusion 2. Across all datasets, we only observe a small performance drop when using $\Delta_\text{Min}$, instead of $\Delta_{\text{AE}_i}$ with the matching AE, which is crucial for applying \method{} in real-world scenarios. As a sanity check, we determine for each sample which AE provides the most accurate reconstruction, i.e., for which AE the reconstruction error $\Delta_{\text{AE}_i}$ is equal to the minimum reconstruction error $\Delta_\text{Min}$.
The results in \cref{tab:attribution_conv2} do not only confirm the assumption that the correct AE achieves the lowest reconstruction error, but also demonstrate that \method{} can be used to attribute images to a specific generative model. However, attribution is limited by the fact that some generative models use the same AE, e.g., Stable Diffusion 1.1 and 1.5.

\subsection{Comparison to Baselines}\label{sec:experiments:comparison}

\begin{table*}
    \setlength{\tabcolsep}{4.0pt}
    \centering
    \scriptsize
    \begin{tabular}{@{\ }lrrrrrrrrrrrrrrr@{\ }}
        \toprule
        & & \multicolumn{7}{c}{AP} & \multicolumn{7}{c}{TPR@5\%FPR} \\
         \cline(l{.3em}r{.3em}){3-9} \cline(l{.3em}r{.3em}){10-16}
        & Training Samples & SD1.1 & SD1.5 & SD2.1 & KD2.1 & MJ4 & MJ5 & MJ5.1 & SD1.1 & SD1.5 & SD2.1 & KD2.1 & MJ4 & MJ5 & MJ5.1 \\
        \midrule
        Gragnaniello~\citep{gragnanielloAreGANGenerated2021} & \num[mode=text]{720000} & 0.715 & 0.701 & 0.629 & 0.526 & 0.664 & 0.666 & 0.598 & 0.149 & 0.151 & 0.107 & 0.043 & 0.160 & 0.163 & 0.108 \\
        Corvi~\citep{corviDetectionSyntheticImages2023} & \num[mode=text]{400000} & \textbf{1.000} & \textbf{1.000} & \textbf{1.000} & \textbf{1.000} & \textbf{1.000} & \textbf{1.000} & \textbf{1.000} & \textbf{1.000} & \textbf{1.000} & \textbf{1.000} & \textbf{0.999} & \textbf{1.000} & \textbf{1.000} & \textbf{0.999} \\
        Ohja~\citep{ojhaUniversalFakeImage2023} & \num[mode=text]{720000} & 0.895 & 0.835 & 0.732 & 0.744 & 0.756 & 0.713 & 0.682 & 0.596 & 0.416 & 0.256 & 0.287 & 0.274 & 0.236 & 0.205 \\
        DIRE~\citep{wangDIREDiffusiongeneratedImage2023} & \num[mode=text]{80000} & 0.457 & 0.457 & 0.480 & 0.513 & 0.503 & 0.498 & 0.500 & 0.000 & 0.000 & 0.000 & 0.000 & 0.000 & 0.000 & 0.000 \\
        \midrule
        $\text{SeDID}_\text{Stat}$~\citep{maexposing2023} & - & 0.484 & 0.783 & 0.713 & 0.728 & 0.376 & 0.401 & 0.403 & 0.049 & 0.308 & 0.205 & 0.259 & 0.005 & 0.011 & 0.012 \\
        \textbf{\method{}}  & - & \textbf{0.979} & \textbf{0.978} & \textbf{0.994} & \textbf{0.999} & \textbf{0.999} & \textbf{0.997} & \textbf{0.996} & \textbf{0.981} & \textbf{0.963} & \textbf{0.995} & \textbf{0.999} & \textbf{0.997} & \textbf{0.993} & \textbf{0.989} \\
        \bottomrule
    \end{tabular}
    \caption{Detection performance of \method{} and baselines, measured in AP and TPR@5\%FPR. All methods in the upper section require training, while those in the lower section are training-free. The results for our method are obtained with $\text{LPIPS}_\text{2}$ and $\Delta_\text{Min}$ (corresponding to the 12\textsuperscript{th} row in \cref{tab:lpips_layers}). The best result for each dataset (per section) is highlighted in \textbf{bold}.}
    \label{tab:baselines}
\end{table*}

We compare \method{} against several state-of-art baselines (see \cref{sec:related_work}) using code and pre-trained models provided by the original authors. An exception is $\text{SeDID}_\text{Stat}$~\citep{maexposing2023}, which has only been evaluated on conventional DMs. To allow for a fair comparison, we adapt their approach to text-to-image LDMs. We also allow $\text{SeDID}_\text{Stat}$ to ``cheat'', by using the matching model for each dataset (except for Midjourney, for which we use SD2.1). We provide more details on how we evaluate the baselines in \cref{sup:baselines} in the supplementary material.

The results in \cref{tab:baselines} show the AP and the true positive rate (TPR) at a fixed false positive rate (FPR). The latter is better suited to estimate the usefulness of a detector in realistic settings, in which only a certain FPR (\SI{5}{\percent} in our case) is tolerable. Among all methods, only \method{} and the method by \citet{corviDetectionSyntheticImages2023} can reliably detect generated images from all models. Despite involving no training, our method achieves almost the same performance as the deep classifier directly trained on generated images. The only other training-free approach, $\text{SeDID}_\text{Stat}$~\citep{maexposing2023}, achieves mediocre performance for some generative models, but fails completely to detect images from other generators. The universal detection approach by \citet{ojhaUniversalFakeImage2023} performs decent on most models, especially SD1.1 and 1.5, but by taking the more realistic TPR@5\%FPR metric into account, its deficiencies become apparent. We suppose the performance drop, in comparison to the results in the original publication, stem from the fact that the classifier was trained on smaller images ($256\times256$). Surprisingly, we obtain very poor results with DIRE~\citep{wangDIREDiffusiongeneratedImage2023}, which contradicts the results reported by the authors, who claim that their method generalizes to images generated by Stable Diffusion 1.5. We took great care to use their public code and pre-trained models as specified in the documentation.
In our analysis, almost all images (both real and generated) were classified as being generated. Upon close inspection of code, data, and models provided by the authors, we suspect that the DIRE representations used to train the classifiers were saved inconsistently. In particular, the DIREs of real images were compressed using JPEG while the DIREs of generated images were stored in lossless PNG format. This unwanted bias would explain why all of our uncompressed DIREs are classified as being generated. We provide a thorough analysis in \cref{sec:dire_analysis} in the supplementary material.

\subsection{Qualitative Image Analysis with \method{}}\label{sec:experiments:qualitative}
In this section we first examine how the reconstruction error of real and generated images is influenced by image complexity. We then illustrate how \method{} can be leveraged to identify inpainted regions within real images.

\paragraph{Relation Between Image Complexity and Reconstruction Error}

\def\eightcolsfactor{0.052}

\begin{figure}
    \centering
    \begin{subfigure}{\eightcolsfactor\textwidth}
        \includegraphics[width=\linewidth]{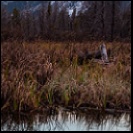}
    \end{subfigure}
    \begin{subfigure}{\eightcolsfactor\textwidth}
        \includegraphics[width=\linewidth]{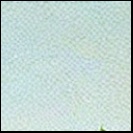}
    \end{subfigure}
    \begin{subfigure}{\eightcolsfactor\textwidth}
        \includegraphics[width=\linewidth]{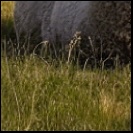}
    \end{subfigure}
    \begin{subfigure}{\eightcolsfactor\textwidth}
        \includegraphics[width=\linewidth]{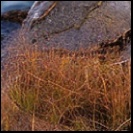}
    \end{subfigure}
    \begin{subfigure}{\eightcolsfactor\textwidth}
        \includegraphics[width=\linewidth]{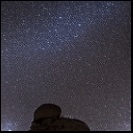}
    \end{subfigure}
    \begin{subfigure}{\eightcolsfactor\textwidth}
        \includegraphics[width=\linewidth]{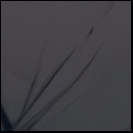}
    \end{subfigure}
    \begin{subfigure}{\eightcolsfactor\textwidth}
        \includegraphics[width=\linewidth]{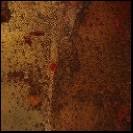}
    \end{subfigure}
    \begin{subfigure}{\eightcolsfactor\textwidth}
        \includegraphics[width=\linewidth]{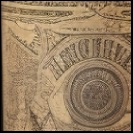}
    \end{subfigure}
    \hfill
    \begin{subfigure}{\eightcolsfactor\textwidth}
        \includegraphics[width=\linewidth]{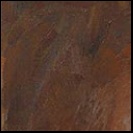}
    \end{subfigure}
    \begin{subfigure}{\eightcolsfactor\textwidth}
        \includegraphics[width=\linewidth]{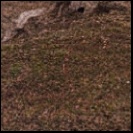}
    \end{subfigure}
    \begin{subfigure}{\eightcolsfactor\textwidth}
        \includegraphics[width=\linewidth]{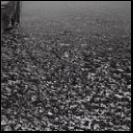}
    \end{subfigure}
    \begin{subfigure}{\eightcolsfactor\textwidth}
        \includegraphics[width=\linewidth]{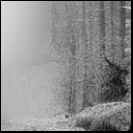}
    \end{subfigure}
    \begin{subfigure}{\eightcolsfactor\textwidth}
        \includegraphics[width=\linewidth]{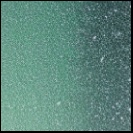}
    \end{subfigure}
    \begin{subfigure}{\eightcolsfactor\textwidth}
        \includegraphics[width=\linewidth]{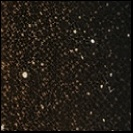}
    \end{subfigure}
    \begin{subfigure}{\eightcolsfactor\textwidth}
        \includegraphics[width=\linewidth]{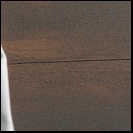}
    \end{subfigure}
    \begin{subfigure}{\eightcolsfactor\textwidth}
        \includegraphics[width=\linewidth]{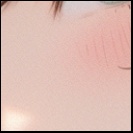}
    \end{subfigure}
    \hfill
    \vspace*{0.1cm}%
    \begin{subfigure}{\eightcolsfactor\textwidth}
        \includegraphics[width=\linewidth]{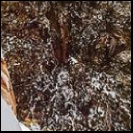}
    \end{subfigure}
    \begin{subfigure}{\eightcolsfactor\textwidth}
        \includegraphics[width=\linewidth]{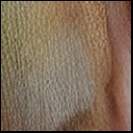}
    \end{subfigure}
    \begin{subfigure}{\eightcolsfactor\textwidth}
        \includegraphics[width=\linewidth]{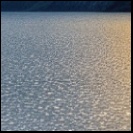}
    \end{subfigure}
    \begin{subfigure}{\eightcolsfactor\textwidth}
        \includegraphics[width=\linewidth]{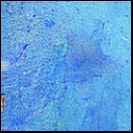}
    \end{subfigure}
    \begin{subfigure}{\eightcolsfactor\textwidth}
        \includegraphics[width=\linewidth]{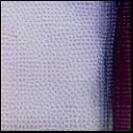}
    \end{subfigure}
    \begin{subfigure}{\eightcolsfactor\textwidth}
        \includegraphics[width=\linewidth]{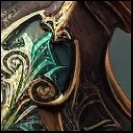}
    \end{subfigure}
    \begin{subfigure}{\eightcolsfactor\textwidth}
        \includegraphics[width=\linewidth]{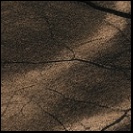}
    \end{subfigure}
    \begin{subfigure}{\eightcolsfactor\textwidth}
        \includegraphics[width=\linewidth]{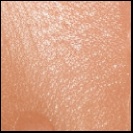}
    \end{subfigure}
    \hfill
    \begin{subfigure}{\eightcolsfactor\textwidth}
        \includegraphics[width=\linewidth]{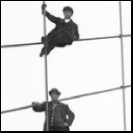}
    \end{subfigure}
    \begin{subfigure}{\eightcolsfactor\textwidth}
        \includegraphics[width=\linewidth]{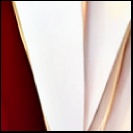}
    \end{subfigure}
    \begin{subfigure}{\eightcolsfactor\textwidth}
        \includegraphics[width=\linewidth]{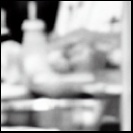}
    \end{subfigure}
    \begin{subfigure}{\eightcolsfactor\textwidth}
        \includegraphics[width=\linewidth]{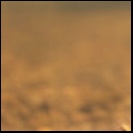}
    \end{subfigure}
    \begin{subfigure}{\eightcolsfactor\textwidth}
        \includegraphics[width=\linewidth]{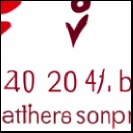}
    \end{subfigure}
    \begin{subfigure}{\eightcolsfactor\textwidth}
        \includegraphics[width=\linewidth]{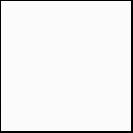}
    \end{subfigure}
    \begin{subfigure}{\eightcolsfactor\textwidth}
        \includegraphics[width=\linewidth]{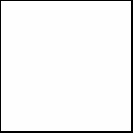}
    \end{subfigure}
    \begin{subfigure}{\eightcolsfactor\textwidth}
        \includegraphics[width=\linewidth]{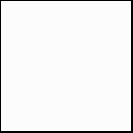}
    \end{subfigure}
    \hfill
    \begin{subfigure}{\eightcolsfactor\textwidth}
        \includegraphics[width=\linewidth]{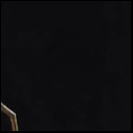}
    \end{subfigure}
    \begin{subfigure}{\eightcolsfactor\textwidth}
        \includegraphics[width=\linewidth]{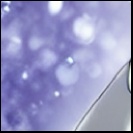}
    \end{subfigure}
    \begin{subfigure}{\eightcolsfactor\textwidth}
        \includegraphics[width=\linewidth]{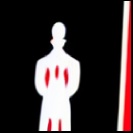}
    \end{subfigure}
    \begin{subfigure}{\eightcolsfactor\textwidth}
        \includegraphics[width=\linewidth]{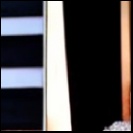}
    \end{subfigure}
    \begin{subfigure}{\eightcolsfactor\textwidth}
        \includegraphics[width=\linewidth]{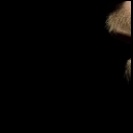}
    \end{subfigure}
    \begin{subfigure}{\eightcolsfactor\textwidth}
        \includegraphics[width=\linewidth]{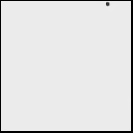}
    \end{subfigure}
    \begin{subfigure}{\eightcolsfactor\textwidth}
        \includegraphics[width=\linewidth]{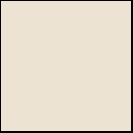}
    \end{subfigure}
    \begin{subfigure}{\eightcolsfactor\textwidth}
        \includegraphics[width=\linewidth]{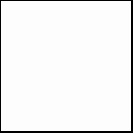}
    \end{subfigure}
    \hfill
    \begin{subfigure}{\eightcolsfactor\textwidth}
        \includegraphics[width=\linewidth]{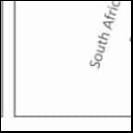}
        \caption*{Real}
    \end{subfigure}
    \begin{subfigure}{\eightcolsfactor\textwidth}
        \includegraphics[width=\linewidth]{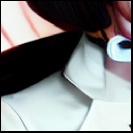}
        \caption*{SD1.1}
    \end{subfigure}
    \begin{subfigure}{\eightcolsfactor\textwidth}
        \includegraphics[width=\linewidth]{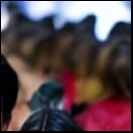}
        \caption*{SD1.5}
    \end{subfigure}
    \begin{subfigure}{\eightcolsfactor\textwidth}
        \includegraphics[width=\linewidth]{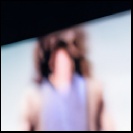}
        \caption*{SD2.1}
    \end{subfigure}
    \begin{subfigure}{\eightcolsfactor\textwidth}
        \includegraphics[width=\linewidth]{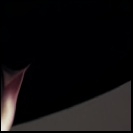}
        \caption*{KD2.1}
    \end{subfigure}
    \begin{subfigure}{\eightcolsfactor\textwidth}
        \includegraphics[width=\linewidth]{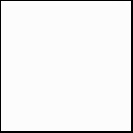}
        \caption*{MJ4}
    \end{subfigure}
    \begin{subfigure}{\eightcolsfactor\textwidth}
        \includegraphics[width=\linewidth]{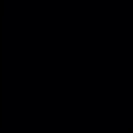}
        \caption*{MJ5}
    \end{subfigure}
    \begin{subfigure}{\eightcolsfactor\textwidth}
        \includegraphics[width=\linewidth]{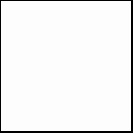}
        \caption*{MJ5.1}
    \end{subfigure}
    \caption{Example patches of size $128\times128$ with high (upper half) and low (lower half) reconstruction error. The reconstruction error is computed using $\text{LPIPS}_2$ and $\Delta_\text{Min}$, and the patches are selected from the top and bottom \SI{1}{\percent} of each dataset.}
    \label{fig:lpips_features}
\end{figure}

\begin{figure}
    \centering
    \begin{subfigure}{0.49\linewidth}
        \includegraphics{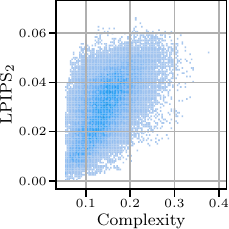}
        \caption{Real}
        \label{fig:lpips_vs_complexity:real}
    \end{subfigure}
    \begin{subfigure}{0.49\linewidth}
        \includegraphics{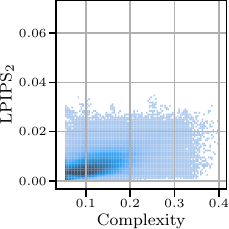}
        \caption{Generated}
    \end{subfigure}
    \caption{Reconstruction error against complexity for real and generated images. (b) contains generated images from all datasets, we provide individual plots in \cref{sup:complexity} in the supplementary material. The color map is clipped at \num{1000} samples for better visibility.}
    \label{fig:lpips_vs_complexity}
\end{figure}

During our experiments we make the observation that simple parts of images, like monochrome areas, can be reconstructed more accurately than complex parts. The examples in \cref{fig:lpips_features} illustrate this behavior. However, we find that the relation between reconstruction error and image complexity differs between real and generated images. In \cref{fig:lpips_vs_complexity} we plot $\Delta_\text{Min}$ against the image complexity for overlapping patches of size $128\times128$ with stride $64$. We estimate the complexity of a patch by its file size after JPEG compression with quality $50$, which approximates the Kolmogorov complexity~\citep{yuImageComplexitySpatial2013,cilibrasiClusteringCompression2005}. For real images, we observe that the reconstruction error is positively correlated with image complexity. For generated images, this trend is significantly less pronounced. In particular, patches with high complexity can be reconstructed considerably better compared to real images. These results suggest that if the fine-grained details of an image can be accurately reconstructed, it is likely generated by an LDM. This allows for a qualitative image analysis, which can explain \method{}'s predictions. In the next section we demonstrate how this analysis can help to localize inpainted regions. It should be noted that this property makes generated images with low complexity, e.g., logos, harder to detect with our method. However, we argue that these are less harmful than complex, photorealistic images.

\def\fivecolsfactor{0.18}

\begin{figure}
    \centering
    \begin{subfigure}{\fivecolsfactor\columnwidth}
        \includegraphics[width=\linewidth]{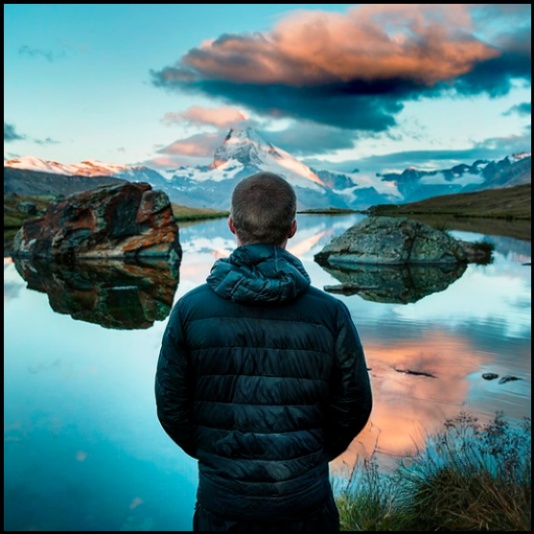}
    \end{subfigure}
    \begin{subfigure}{\fivecolsfactor\columnwidth}
        \includegraphics[width=\linewidth]{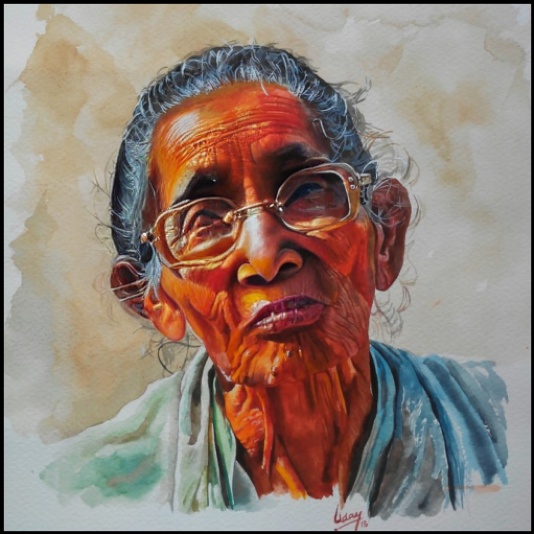}
    \end{subfigure}
    \begin{subfigure}{\fivecolsfactor\columnwidth}
        \includegraphics[width=\linewidth]{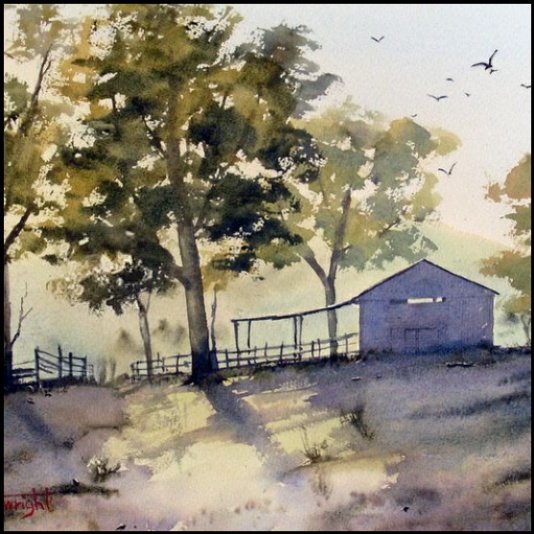}
    \end{subfigure}
    \begin{subfigure}{\fivecolsfactor\columnwidth}
        \includegraphics[width=\linewidth]{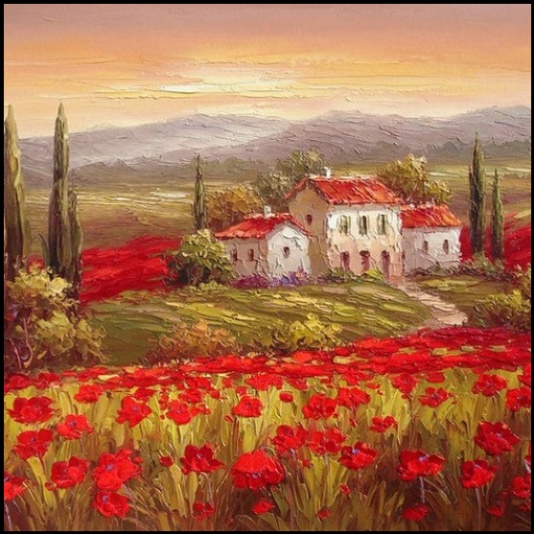}
    \end{subfigure}
    \begin{subfigure}{\fivecolsfactor\columnwidth}
        \includegraphics[width=\linewidth]{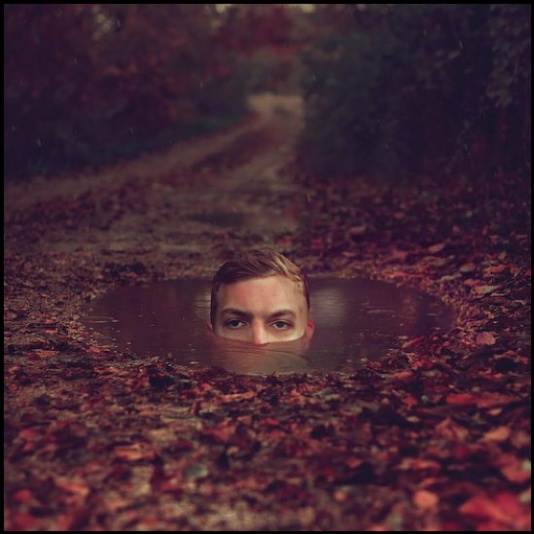}
    \end{subfigure}
    \hfill
    \begin{subfigure}{\fivecolsfactor\columnwidth}
        \includegraphics[width=\linewidth]{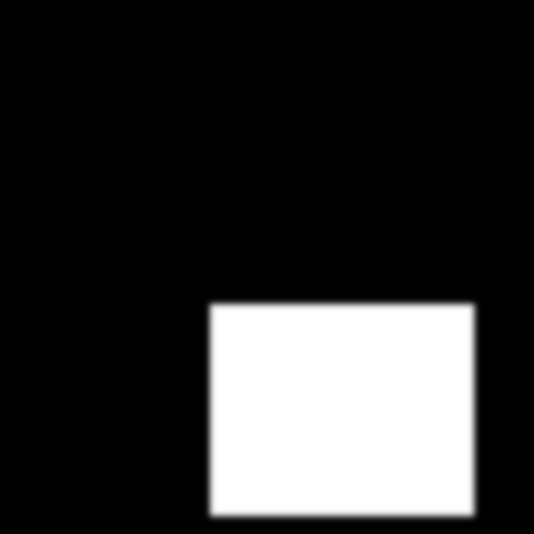}
    \end{subfigure}
    \begin{subfigure}{\fivecolsfactor\columnwidth}
        \includegraphics[width=\linewidth]{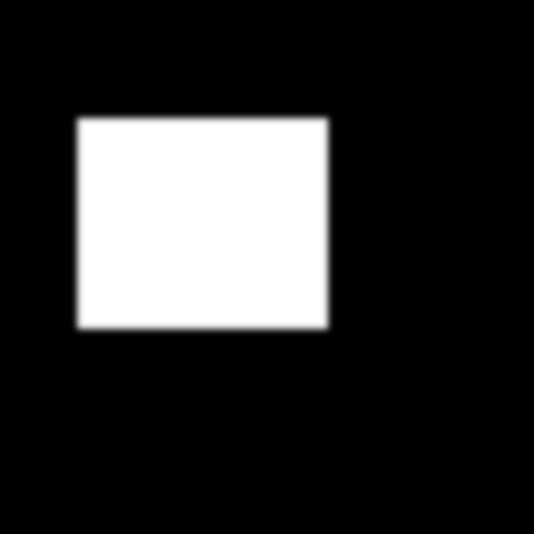}
    \end{subfigure}
    \begin{subfigure}{\fivecolsfactor\columnwidth}
        \includegraphics[width=\linewidth]{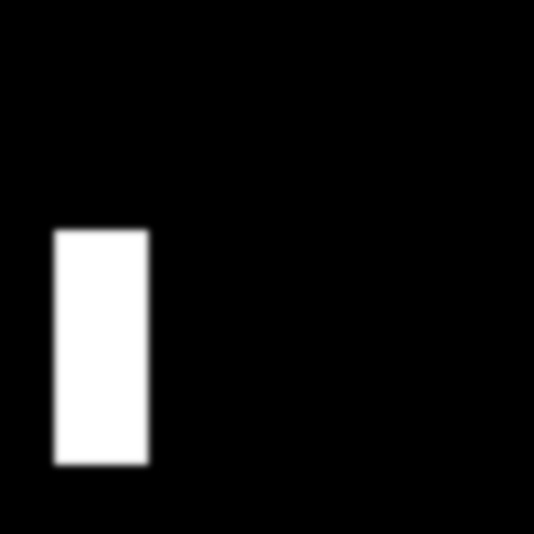}
    \end{subfigure}
    \begin{subfigure}{\fivecolsfactor\columnwidth}
        \includegraphics[width=\linewidth]{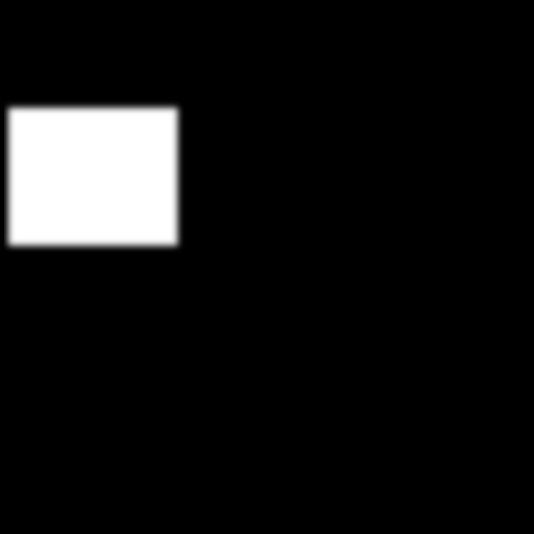}
    \end{subfigure}
    \begin{subfigure}{\fivecolsfactor\columnwidth}
        \includegraphics[width=\linewidth]{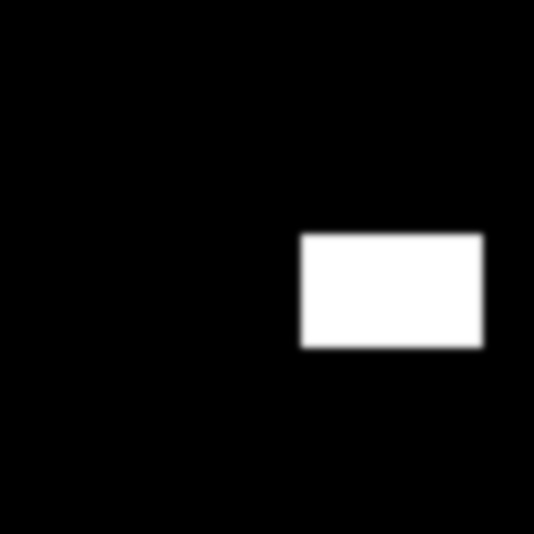}
    \end{subfigure}
    \hfill
    \begin{subfigure}{\fivecolsfactor\columnwidth}
        \includegraphics[width=\linewidth]{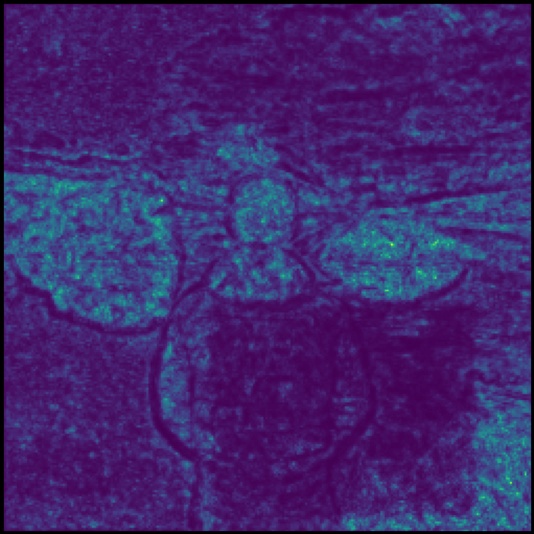}
    \end{subfigure}
    \begin{subfigure}{\fivecolsfactor\columnwidth}
        \includegraphics[width=\linewidth]{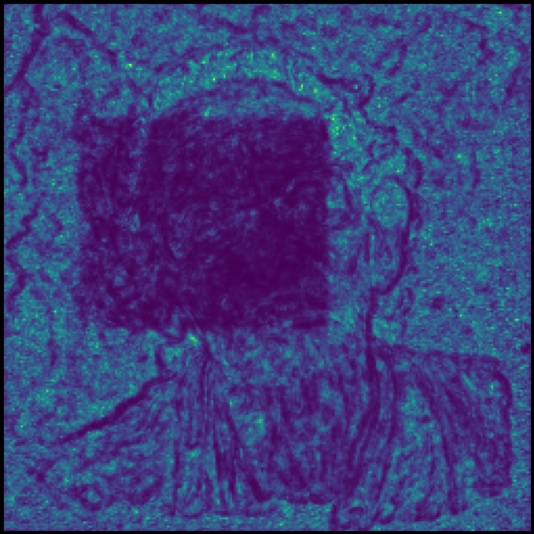}
    \end{subfigure}
    \begin{subfigure}{\fivecolsfactor\columnwidth}
        \includegraphics[width=\linewidth]{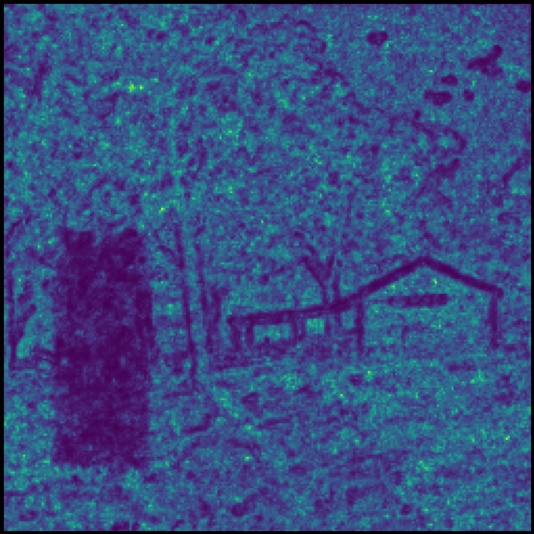}
    \end{subfigure}
    \begin{subfigure}{\fivecolsfactor\columnwidth}
        \includegraphics[width=\linewidth]{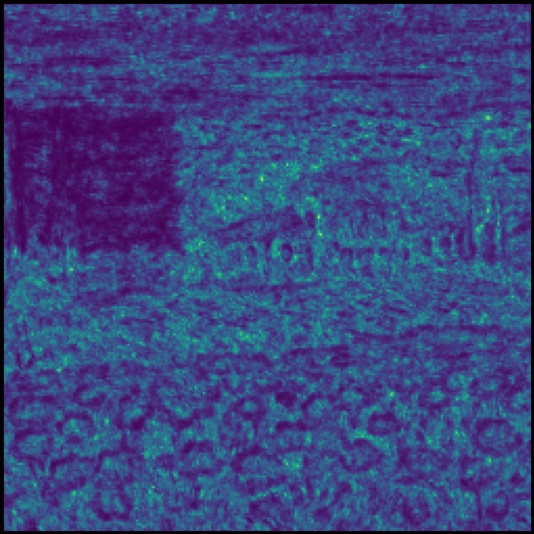}
    \end{subfigure}
    \begin{subfigure}{\fivecolsfactor\columnwidth}
        \includegraphics[width=\linewidth]{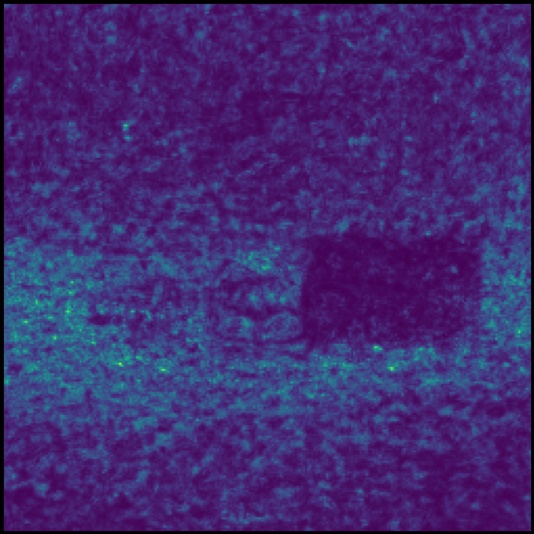}
    \end{subfigure}
    \caption{Examples illustrating the localization of inpainted regions using the reconstruction error. We show the images inpainted with Stable Diffusion 1.5 (top), the masks used for inpainting (center) and the reconstruction error maps (bottom), computed with $\text{LPIPS}_2$ and the AE from Stable Diffusion 1.5. We provide more examples in \cref{sup:inpainting} in the supplementary material.}
    \label{fig:inpainting_examples}
\end{figure}

\paragraph{Localizing Inpainted Regions}
In \cref{fig:inpainting_examples} we show that inspecting the reconstruction error \textit{map} of an image can provide hints to identify inpainted regions within an authentic image.
To illustrate this, we take real images and inpaint a randomly generated rectangular mask using the inpainting variant of Stable Diffusion 1.5.
We then compute $\Delta_{\text{AE}_i}$ with the AE from Stable Diffusion 1 but omit spatial averaging to obtain the heatmaps shown in the third row of \cref{fig:inpainting_examples}.
The inpainted regions can be identified due to their significantly lower reconstruction error, which is clearly noticeable in regions with higher complexity.
In future work we aim to investigate how the reconstruction error can be used to predict the precise locations of inpainted regions.

\subsection{Additional Analyses}\label{sec:experiments:additional}
Finally, we conduct additional experiments to better understand the properties of \method{}. We analyze its robustness to common image perturbations, evaluate the influence of the used distance metric, and explore computing the errors from deeper reconstructions.

\begin{figure}
    \centering
    \begin{subfigure}{0.49\linewidth}
        \centering
        \includegraphics{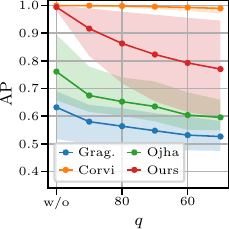}
        \caption{JPEG}
    \end{subfigure}
    \begin{subfigure}{0.49\linewidth}
        \centering
        \includegraphics{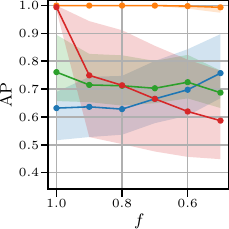}
        \caption{Crop}
    \end{subfigure}
    \begin{subfigure}{0.49\linewidth}
        \centering
        \includegraphics{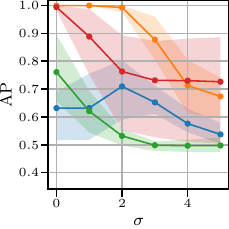}
        \caption{Blur}
    \end{subfigure}
    \begin{subfigure}{0.49\linewidth}
        \centering
        \includegraphics{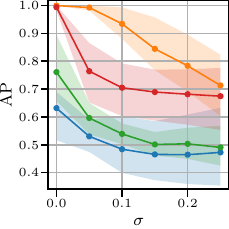}
        \caption{Noise}
    \end{subfigure}
    \caption{Detection performance of \method{} (with $\text{LPIPS}_\text{2}$ and $\Delta_\text{Min}$) and baselines on perturbed images, measured in AP. Results are averaged over all datasets, with shaded areas indicating the minimum and maximum. We provide extended results in \cref{sup:robustness} in the supplementary material.}
    \label{fig:perturbations_ap}
\end{figure}

\paragraph{Robustness to Perturbations}
In real-world scenarios, images are often processed, e.g., during the upload to social media platforms, which may affect detection performance.
We therefore evaluate how robust our method is to common image perturbations. Following previous works~\citep{frankLeveragingFrequencyAnalysis2020,yuResponsibleDisclosureGenerative2022} we use JPEG compression (with quality $q$), center cropping (with crop factor $f$ and subsequent resizing to the original size), Gaussian blur, and Gaussian noise (both with standard deviation $\sigma$).
In \cref{fig:perturbations_ap} we report the performance of \method{} and several baselines on a set of $250$ real and $250$ generated images (we omit DIRE~\citep{wangDIREDiffusiongeneratedImage2023} and $\text{SeDID}_\text{Stat}$~\citep{maexposing2023} due to the low performance on clean images and high computational cost).
We find that our method outperforms the detectors from \citet{gragnanielloAreGANGenerated2021} and \citet{ojhaUniversalFakeImage2023} in most settings. The results for individual datasets and layers (see~\cref{sup:robustness} in the supplementary material) also reveal that the robustness strongly depends on the dataset and selected LPIPS layer. For instance, $\text{LPIPS}_4$ appears to be significantly less affected by perturbations, possibly due to the larger receptive field. While the method proposed by \citet{corviDetectionSyntheticImages2023} is more robust, we emphasize that it was directly trained on perturbed, LDM-generated images. Our method, in contrast, is completely training-free.
In future work, we plan to explore whether using a more robust distance metric or adding a simple classifier trained on the reconstruction errors can increase the robustness of \method{}.

\paragraph{Exploring Different Distance Metrics}
In \cref{tab:distanc_metrics} we compare how the distance metric influences the performance of \method{}. Beside VGG16~\citep{simonyanVeryDeepConvolutional2015}, which we use in previous experiments, we consider SqueezeNet~\citep{iandolaSqueezeNetAlexNetlevelAccuracy2016} and AlexNet~\citep{krizhevskyImageNetClassificationDeep2012} as backbones for LPIPS (using its standard definition). Moreover, we test alternative similarity metrics like mean squared error (MSE), structural similarity index (SSIM)~\citep{wangImageQualityAssessment2004a}, multiscale structural similarity index (MS-SSIM)~\citep{wangMultiscaleStructuralSimilarity2003}, and DISTS~\citep{dingImageQualityAssessment2022}. The results show that LPIPS~(VGG16) achieves the best overall performance, while other metrics achieve high AP only on some datasets.

\begin{table}
    \setlength{\tabcolsep}{4.0pt}
    \centering
    \scriptsize
    \begin{tabular}{@{\ }l@{\hspace{6pt}}rrrrrrr@{\ }}
        \toprule
        Distance & SD1.1 & SD1.5 & SD2.1 & KD2.1 & MJ4 & MJ5 & MJ5.1 \\
        \midrule
        LPIPS (VGG16) & \textbf{0.959} & \textbf{0.957} & \textbf{0.991} & \textbf{0.996} & \textbf{0.998} & \textbf{0.993} & 0.994 \\
        LPIPS (AlexNet) & 0.755 & 0.765 & 0.904 & 0.960 & 0.981 & 0.953 & 0.953 \\
        LPIPS (SqueezeNet) & 0.848 & 0.847 & 0.950 & 0.976 & 0.992 & 0.973 & 0.974 \\
        DISTS~\citep{dingImageQualityAssessment2022} & 0.866 & 0.860 & 0.962 & 0.928 & \textbf{0.998} & 0.991 & \textbf{0.995} \\
        MSE & 0.534 & 0.571 & 0.792 & 0.882 & 0.886 & 0.863 & 0.866 \\
        SSIM~\citep{wangImageQualityAssessment2004a} & 0.808 & 0.812 & 0.924 & 0.981 & 0.984 & 0.957 & 0.955 \\
        MS-SSIM~\citep{wangMultiscaleStructuralSimilarity2003} & 0.842 & 0.854 & 0.955 & 0.984 & 0.989 & 0.977 & 0.977 \\
        \bottomrule
    \end{tabular}
    \caption{Detection performance of \method{} using different distance metrics, measured in AP.}
    \label{tab:distanc_metrics}
\end{table}

\begin{figure}
    \centering
    \begin{subfigure}{\linewidth}
        \includegraphics{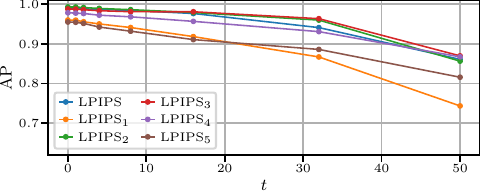}
        \caption{Stable Diffusion 1.5}
    \end{subfigure}
    \begin{subfigure}{\linewidth}
        \includegraphics{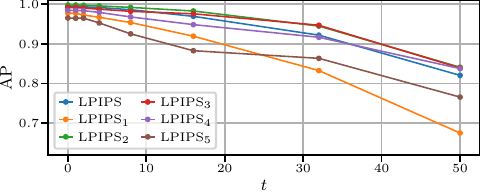}
        \caption{Stable Diffusion 2.1}
    \end{subfigure}
    \caption{Detection performance using deeper reconstructions, measured in AP. Reconstructions are computed using the corresponding AE and $t$ steps of the DM's denoising process.}
    \label{fig:deeper}
\end{figure}

\paragraph{Using Deeper Reconstructions}
Finally, we experiment with reconstructing images using not only the AE, but also (parts of) the denoising process in latent space to study the effect of deeper reconstructions on training-free detection performance.
Here, we first compute latent (noisy) images that would generate the given image by inverting the DDIM sampler~\citep{songDenoisingDiffusionImplicit2022} for several steps and then reconstruct the original image by denoising with the DDIM sampler as usual.
The hypothesis is again that generated images should be reconstructed more accurately using DDIM inversion and denoising.
We set the total number of steps to $50$ and guide the deterministic denoising process by a prompt extracted using BLIP~\citep{liAlignFuseVision2021}. \cref{fig:deeper} depicts the detection performance for Stable Diffusion 1.5 and 2.1 using different amounts of reconstruction steps (out of $50$) based on $250$ samples per dataset. Note that we use the matching AE and U-Net weights for both datasets, respectively, and that $t=0$ corresponds to our previously described approach using just $\Delta_{\text{AE}_i}$.
With one or two steps, the AP is almost equal to the setting with $t=0$ across both datasets and all variants of LPIPS. Increasing the number of steps causes a notable decrease in detection performance, especially for higher LPIPS layers.
Note that the reconstructions obtained with the full denoising process (i.e., $t=50$) correspond to the reconstructions that DIRE \citep{wangDIREDiffusiongeneratedImage2023} uses.
We conclude that including the denoising process to compute the reconstruction distance does not benefit the detection performance, especially given the added computational complexity.
Another disadvantage is that it requires the weights of the corresponding U-Net, which could be more difficult to obtain than just the AE.

\section{Discussion and Conclusion}\label{sec:discussion}
With LDMs being a key enabler for high-resolution image synthesis, their forensic analysis and the development of specialized detection methods is long overdue. In this work, we attempt to close this gap by proposing \method{}, a simple and training-free approach which can reliably detect images from state-of-the-art LDMs like Stable Diffusion and Midjourney. We find that, despite being training-free, our method achieves a detection performance which is comparable to deep classifiers that are directly trained on LDM-generated images, making it a promising alternative.

A limitation of our method is that, to get the best results, access to the AE of the LDM which actually generated the image is required (see \cref{sec:experiments:detection}). However, we argue that this does not substantially restrict the usability of \method{} in real-world settings. The most widely used LDMs, especially Stable Diffusion, have become so popular due to being publicly available. Furthermore, the fact that the open-source community contributes major innovations to existing models~\citep{patelGoogleWeHave2023} results in free and proprietary models sharing the same characteristics. Our observation that images generated by Midjourney can be detected almost perfectly by using the AE from Stable Diffusion 2 is a prime example for this. We are therefore convinced that with a sufficiently large pool of AEs, \method{} is effective against a wide range of practically relevant LDMs. %
The modularity of our approach is actually a key benefit, since extending it to new models is trivial and does not require expensive re-training.

We realize that this property can also be used as a simple means for model inventors to responsibly disclose new generative models. Existing approaches for responsible model disclosure~\citep{yuArtificialFingerprintingGenerative2021,yuResponsibleDisclosureGenerative2022,fernandezStableSignatureRooting2023} usually require changes to either the training data or the model itself, which might be undesirable or even infeasible. With \method{}, model inventors only have to publish their custom AE, whereas the backbone operating in latent space, which typically is the most valuable asset, remains private. By doing so, model inventors can alleviate potential negative consequences of their work with little to no additional overhead.

We hope that our work offers a novel perspective on the detection of images generated by modern text-to-image models. We believe that \method{} and potential follow-up works can contribute to mitigating the threats of these models to our digital society.

\section*{Acknowledgements}
Funded by the Deutsche Forschungsgemeinschaft (DFG, German Research Foundation) under Germany's Excellence Strategy - EXC 2092 CASA - 390781972.

{
    \small
    \bibliographystyle{ieeenat_fullname}
    \bibliography{main}
}

\onecolumn
\clearpage
{
\centering
\Large
\textbf{\thetitle}\\
\vspace{0.5em}Supplementary Material \\
\vspace{1.0em}
}

\section{Implementation Details}
\subsection{Data Collection}\label{sup:data_collection}
\paragraph{Real}
To obtain real images of high visual quality, we resort to LAION-Aesthetics\footnote{\url{https://laion.ai/blog/laion-aesthetics}}, which defines several subsets from LAION-5B~\citep{schuhmannLAION5BOpenLargescale2022} based on image aesthetics. The authors train a linear model on top of CLIP~\citep{radfordLearningTransferableVisual2021} features to predict the aesthetics of an image, which is then used to create multiple collections.
For our experiments we choose images with an aesthetics score of $6.5$ or higher. Since the images come in various resolutions, we only use images whose smaller side has at least $512$ pixels but whose total number of pixels is less or equal to $768^2$. We then take the center crop of size $512\times512$, ensuring that crops contain meaningful content while avoiding resizing operations, which could potentially distort the results.

\paragraph{Stable Diffusion~\citep{stablediffusion} and Kandinsky~\citep{razzhigaevKandinskyImprovedTexttoimage2023}}
We use the Diffusers\footnote{\url{https://github.com/huggingface/diffusers}} library to generate images using the prompts extracted from real images. All images are generated with the default settings and have size $512\times512$. We use the same library to compute the reconstructions based on the AEs.

\paragraph{Midjourney~\citep{midjourney}}
We take images generated by Midjourney from a dataset available on Kaggle\footnote{\url{https://kaggle.com/datasets/iraklip/modjourney-v51-cleaned-data}}. It contains the URLs to images from the official Midjourney Discord server, together with some metadata, including the used version. We filter the dataset by version (v4, v5, and v5.1) and only select images which have size $1024\times1024$.

\subsection{Baselines}\label{sup:baselines}
\paragraph{\citet{gragnanielloAreGANGenerated2021} and \citet{corviDetectionSyntheticImages2023}}
We use the code and model checkpoints from the official repository\footnote{\url{https://github.com/grip-unina/DMimageDetection}} provided by~\citet{corviDetectionSyntheticImages2023}, which also contains the detector from~\citet{gragnanielloAreGANGenerated2021}.

\paragraph{\citet{ojhaUniversalFakeImage2023}}
We use the code and model checkpoints from the official repository\footnote{\url{https://github.com/Yuheng-Li/UniversalFakeDetect}}. According to the code, a center crop of size $224\times224$ is used as input. Since the authors evaluate their method on smaller images ($256\times256$) than we do, we also try resizing images before cropping. However, this does not significantly alter the results.

\paragraph{DIRE~\citep{wangDIREDiffusiongeneratedImage2023}}
We use the code and model checkpoints from the official repository\footnote{\url{https://github.com/ZhendongWang6/DIRE}}. In particular, we compute the DIRE representations using ADM~\citep{dhariwalDiffusionModelsBeat2021} trained on LSUN Bedroom and classify with the corresponding detector. This setting corresponds to the setting in Table 3 in the original paper~\citep{wangDIREDiffusiongeneratedImage2023}, where the authors report good generalization to Stable Diffusion 1.5.

\paragraph{$\text{SeDID}_\text{Stat}$~\citep{maexposing2023}}
At the time of writing, no code is publicly available, which is why we reimplement the method based on the authors' definitions. To guide the denoising process we use the same technique as in our experiment with deeper reconstructions (see \cref{sec:experiments:additional}). We experiment with different values for the total number of steps and $T_{SE}$ and observe that $50$ steps in total and $T_{SE} = 25$ achieves the best results.

\clearpage
\section{Additional Results}

\subsection{Reconstruction Error Histograms}\label{sup:histograms}
\begin{figure}[h]
    \centering
    \begin{subfigure}{0.49\columnwidth}
        \includegraphics{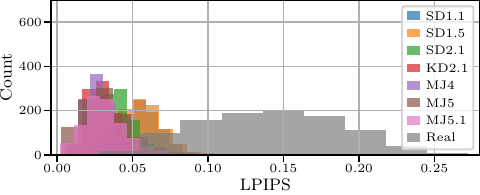}
        \caption{LPIPS}
    \end{subfigure}
    \begin{subfigure}{0.49\columnwidth}
        \includegraphics{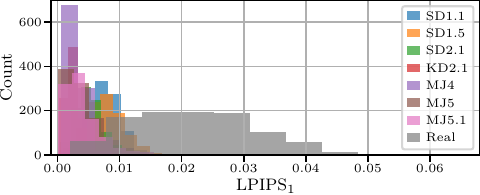}
        \caption{$\text{LPIPS}_1$}
    \end{subfigure}
    \begin{subfigure}{0.49\columnwidth}
        \includegraphics{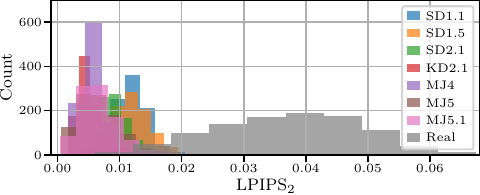}
        \caption{$\text{LPIPS}_2$}
    \end{subfigure}
    \begin{subfigure}{0.49\columnwidth}
        \includegraphics{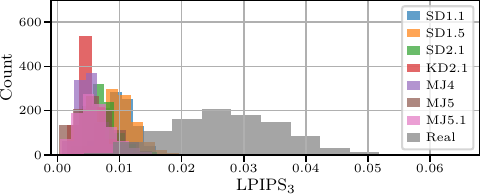}
        \caption{$\text{LPIPS}_3$}
    \end{subfigure}
    \begin{subfigure}{0.49\columnwidth}
        \includegraphics{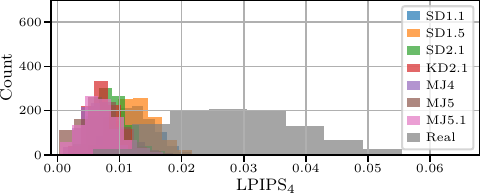}
        \caption{$\text{LPIPS}_4$}
    \end{subfigure}
    \begin{subfigure}{0.49\columnwidth}
        \includegraphics{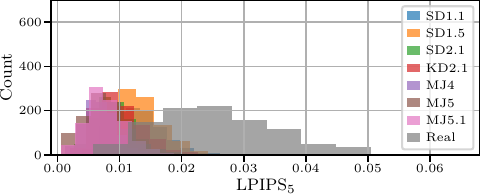}
        \caption{$\text{LPIPS}_5$}
    \end{subfigure}
    \caption{Distributions of reconstruction errors $\Delta_\text{Min}$ using different LPIPS variants. (c) is identical to \cref{fig:hist_conv2} and is included here for completeness. The x-axis in (a) differs because LPIPS is defined as the sum of all layers.}
    \label{fig:histograms}
\end{figure}

\cref{fig:histograms} is the extended version of \cref{fig:hist_conv2} that includes the other variants of LPIPS. For $\text{LPIPS}_2$ the distributions of reconstruction errors for real and generated samples are most separated. This confirms our results in \cref{tab:lpips_layers}, according to which $\text{LPIPS}_2$ achieves the highest APs. We observe that for $\text{LPIPS}_1$, the distributions appear to be shifted to the left, indicating that the reconstruction errors are lower for both real and generated images. For higher layers, the results suggest that the error becomes lower for real images but higher for generated images, making the distributions less separable.

\clearpage
\subsection{Attribution Based on Minimal Reconstruction Error}\label{sup:attribution}
\cref{tab:attribution} is the extended version of \cref{tab:attribution_conv2} that includes all variants of LPIPS. We observe that for LPIPS, $\text{LPIPS}_1$, and $\text{LPIPS}_2$, almost all images are correctly attributed (based on the minimal reconstruction error) to the AE that actually generated them. While this is technically not the case for Midjourney, since the AE is not publicly available, the results strongly indicate that the AE is similar to that of Stable Diffusion 2. Towards higher layers, especially images from Stable Diffusion 1.1 and 1.5 tend to be attributed to the AE of Stable Diffusion 2.

\begin{table}[h]
    \setlength{\tabcolsep}{4.0pt}
    \centering
    \scriptsize
    \begin{tabular}{@{\ }llrrrrrrr@{\ }}
        \toprule
        Distance & AE & SD1.1 & SD1.5 & SD2.1 & KD2.1 & MJ4 & MJ5 & MJ5.1 \\
        \midrule
        \multirow[c]{3}{*}{LPIPS} & SD1 & \textbf{1.000} & 0.999 & 0.000 & 0.000 & 0.000 & 0.000 & 0.000 \\
         & SD2 & 0.000 & 0.001 & \textbf{1.000} & 0.000 & \textbf{0.999} & 0.993 & 0.997 \\
         & KD2.1 & 0.000 & 0.000 & 0.000 & \textbf{1.000} & 0.001 & 0.007 & 0.003 \\
        \cline{1-9}
        \multirow[c]{3}{*}{$\text{LPIPS}_1$} & SD1 & \textbf{1.000} & 0.999 & 0.000 & 0.000 & 0.000 & 0.000 & 0.000 \\
         & SD2 & 0.000 & 0.001 & \textbf{1.000} & 0.000 & \textbf{0.999} & \textbf{0.999} & \textbf{0.999} \\
         & KD2.1 & 0.000 & 0.000 & 0.000 & \textbf{1.000} & 0.001 & 0.001 & 0.001 \\
        \cline{1-9}
        \multirow[c]{3}{*}{$\text{LPIPS}_2$} & SD1 & \textbf{1.000} & \textbf{1.000} & 0.000 & 0.000 & 0.000 & 0.000 & 0.000 \\
         & SD2 & 0.000 & 0.000 & \textbf{1.000} & 0.000 & \textbf{0.999} & 0.997 & 0.995 \\
         & KD2.1 & 0.000 & 0.000 & 0.000 & \textbf{1.000} & 0.001 & 0.003 & 0.005 \\
        \cline{1-9}
        \multirow[c]{3}{*}{$\text{LPIPS}_3$} & SD1 & 0.998 & 0.995 & 0.000 & 0.000 & 0.000 & 0.001 & 0.000 \\
         & SD2 & 0.002 & 0.005 & \textbf{1.000} & 0.000 & 0.997 & 0.991 & 0.995 \\
         & KD2.1 & 0.000 & 0.000 & 0.000 & \textbf{1.000} & 0.003 & 0.008 & 0.005 \\
        \cline{1-9}
        \multirow[c]{3}{*}{$\text{LPIPS}_4$} & SD1 & 0.975 & 0.954 & 0.000 & 0.000 & 0.000 & 0.000 & 0.000 \\
         & SD2 & 0.025 & 0.046 & 0.998 & 0.003 & 0.998 & 0.982 & 0.993 \\
         & KD2.1 & 0.000 & 0.000 & 0.002 & 0.997 & 0.002 & 0.018 & 0.007 \\
        \cline{1-9}
        \multirow[c]{3}{*}{$\text{LPIPS}_5$} & SD1 & 0.835 & 0.772 & 0.001 & 0.001 & 0.000 & 0.000 & 0.000 \\
         & SD2 & 0.162 & 0.228 & 0.995 & 0.013 & 0.997 & 0.977 & 0.990 \\
         & KD2.1 & 0.003 & 0.000 & 0.004 & 0.986 & 0.003 & 0.023 & 0.010 \\
        \bottomrule
        \end{tabular}
    \caption{Fraction of samples for which an AE has the smallest reconstruction error using different LPIPS variants. The highest fraction for each dataset is highlighted in \textbf{bold}. The results for $\text{LPIPS}_2$ are identical to \cref{tab:attribution_conv2} and are included here for completeness.}
    \label{tab:attribution}
\end{table}

\clearpage
\subsection{Reconstruction Error Against Complexity}\label{sup:complexity}
\begin{figure}[h]
    \centering
    \begin{subfigure}{0.245\textwidth}
        \includegraphics{assets/figures/lpips_vs_complexity/Real.pdf}
        \caption{Real}
    \end{subfigure}
    \begin{subfigure}{0.245\textwidth}
        \includegraphics{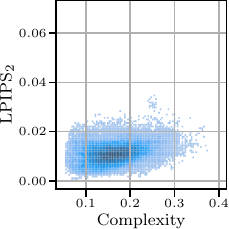}
        \caption{Stable Diffusion 1.1}
    \end{subfigure}
    \begin{subfigure}{0.245\textwidth}
        \includegraphics{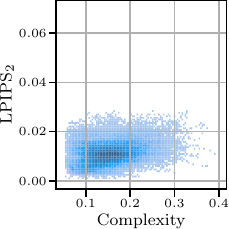}
        \caption{Stable Diffusion 1.5}
    \end{subfigure}
    \begin{subfigure}{0.245\textwidth}
        \includegraphics{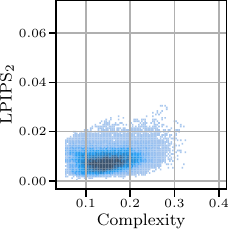}
        \caption{Stable Diffusion 2.1}
    \end{subfigure}
    \hfill
    \begin{subfigure}{0.245\textwidth}
        \includegraphics{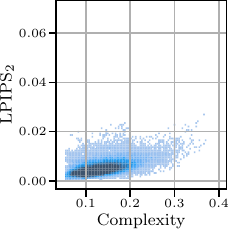}
        \caption{Kandinsky 2.1}
    \end{subfigure}
    \begin{subfigure}{0.245\textwidth}
        \includegraphics{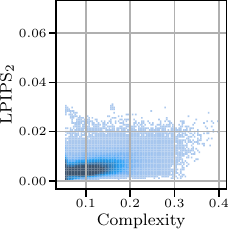}
        \caption{Midjourney v4}
    \end{subfigure}
    \begin{subfigure}{0.245\textwidth}
        \includegraphics{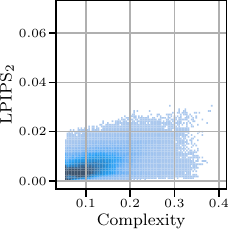}
        \caption{Midjourney v5}
    \end{subfigure}
    \begin{subfigure}{0.245\textwidth}
        \includegraphics{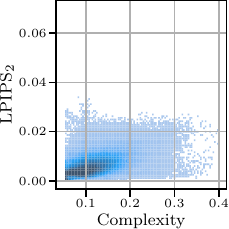}
        \caption{Midjourney v5.1}
    \end{subfigure}
    \caption{Reconstruction error against complexity for different datasets. (a) is identical to \cref{fig:lpips_vs_complexity:real} and is included her for completeness. The color map is clipped at \num{1000} samples for better visibility.}
    \label{fig:lpips_complexity_extended}
\end{figure}

\cref{fig:lpips_complexity_extended} is the extended version of \cref{fig:lpips_vs_complexity} which contains plots for individual datasets. We find that the relation between complexity and reconstruction error is relatively similar for different generative models, with the variants of Stable Diffusion and Kandinsky yielding more compact distributions than Midjourney.

\subsection{Inpainting Localization}\label{sup:inpainting}
In \cref{fig:inpainting_examples_extended} we provide additional examples for reconstruction maps of real images with inpainted regions (corresponding to \cref{fig:inpainting_examples}). Across different scenes, the reconstruction error provides a good indication of where the inpainted region is located.

\def\fivecolsfactor{0.09}

\begin{figure}[h]
    \centering
    \begin{subfigure}{\fivecolsfactor\columnwidth}
        \includegraphics[width=\linewidth]{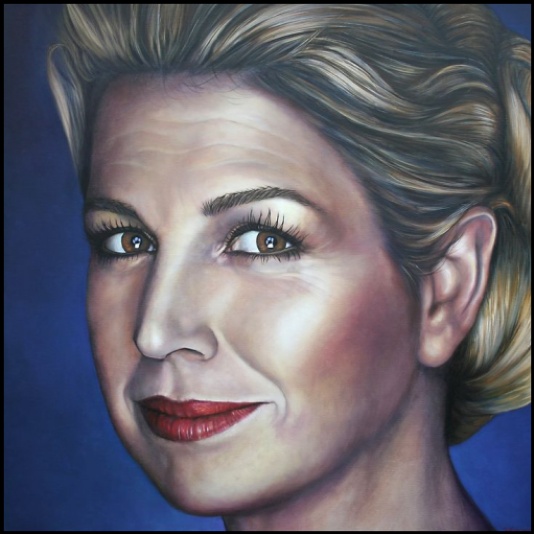}
    \end{subfigure}
    \begin{subfigure}{\fivecolsfactor\columnwidth}
        \includegraphics[width=\linewidth]{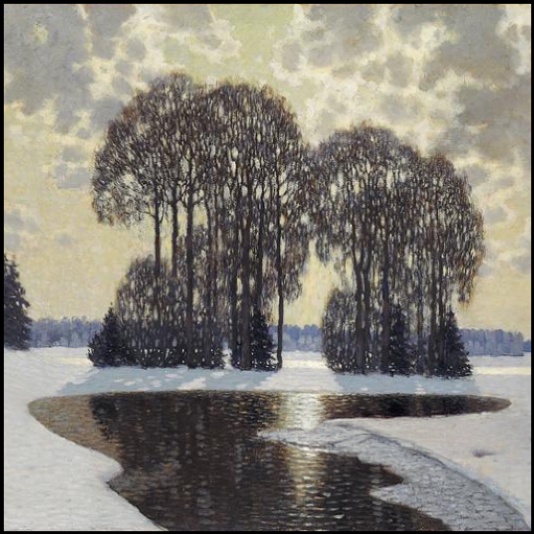}
    \end{subfigure}
    \begin{subfigure}{\fivecolsfactor\columnwidth}
        \includegraphics[width=\linewidth]{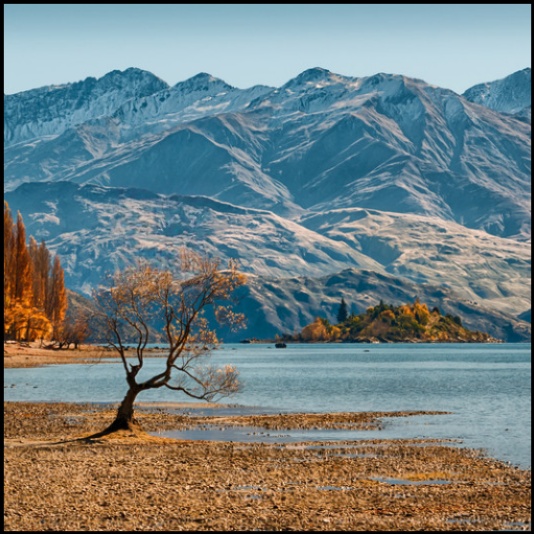}
    \end{subfigure}
    \begin{subfigure}{\fivecolsfactor\columnwidth}
        \includegraphics[width=\linewidth]{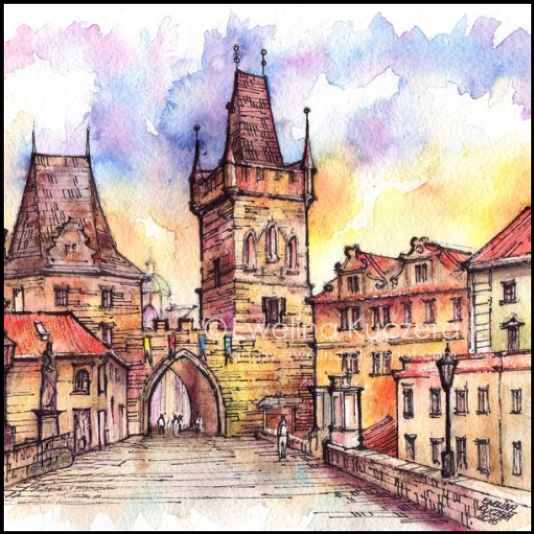}
    \end{subfigure}
    \begin{subfigure}{\fivecolsfactor\columnwidth}
        \includegraphics[width=\linewidth]{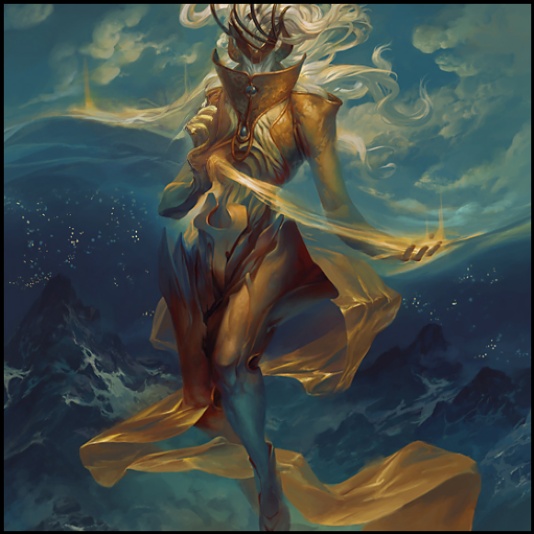}
    \end{subfigure}
    \begin{subfigure}{\fivecolsfactor\columnwidth}
        \includegraphics[width=\linewidth]{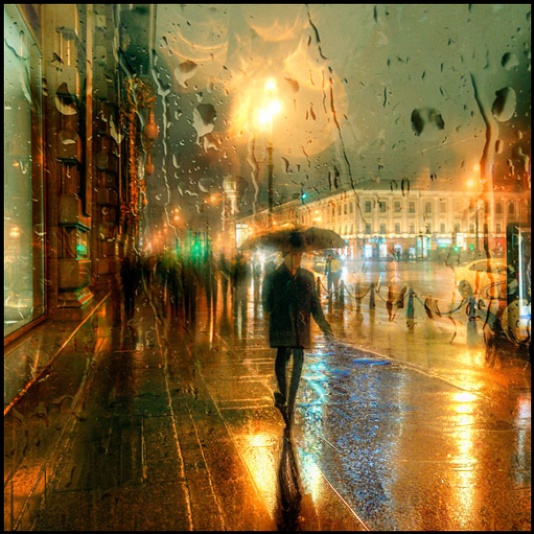}
    \end{subfigure}
    \begin{subfigure}{\fivecolsfactor\columnwidth}
        \includegraphics[width=\linewidth]{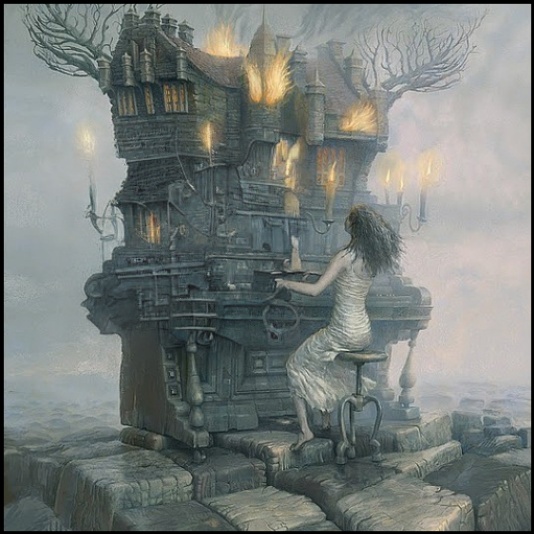}
    \end{subfigure}
    \begin{subfigure}{\fivecolsfactor\columnwidth}
        \includegraphics[width=\linewidth]{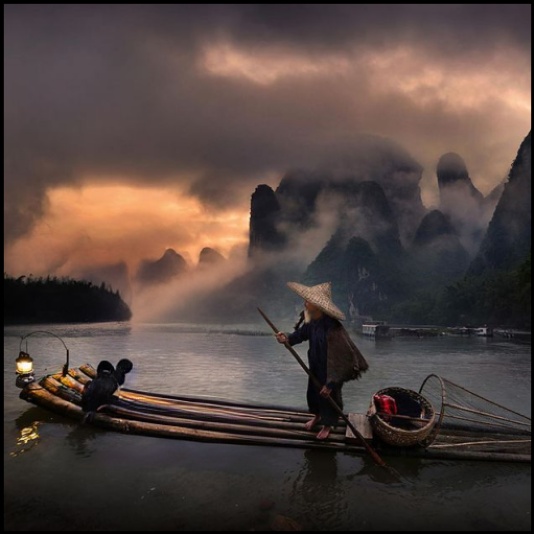}
    \end{subfigure}
    \begin{subfigure}{\fivecolsfactor\columnwidth}
        \includegraphics[width=\linewidth]{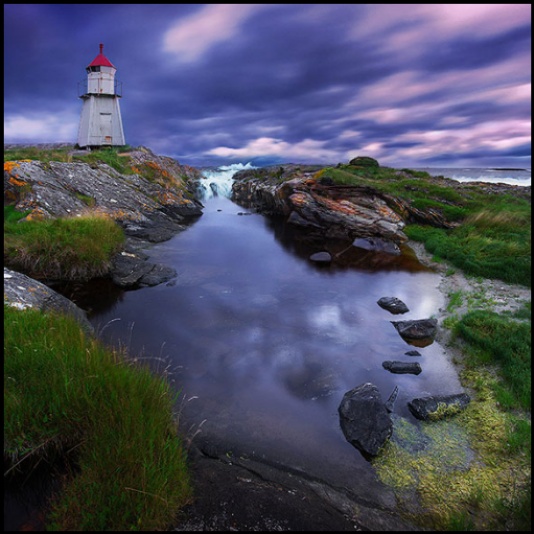}
    \end{subfigure}
    \begin{subfigure}{\fivecolsfactor\columnwidth}
        \includegraphics[width=\linewidth]{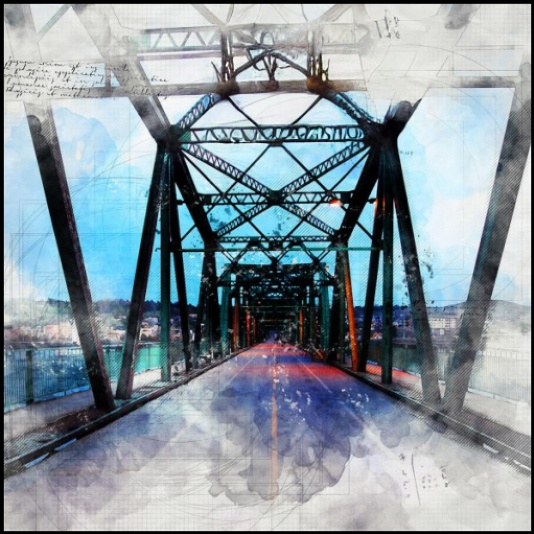}
    \end{subfigure}
    \hfill

    \begin{subfigure}{\fivecolsfactor\columnwidth}
        \includegraphics[width=\linewidth]{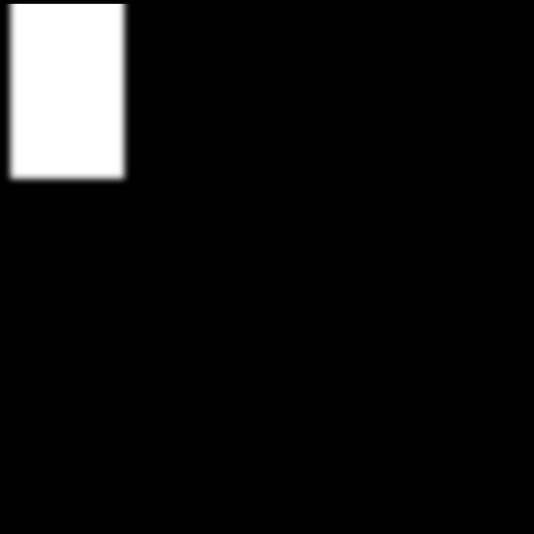}
    \end{subfigure}
    \begin{subfigure}{\fivecolsfactor\columnwidth}
        \includegraphics[width=\linewidth]{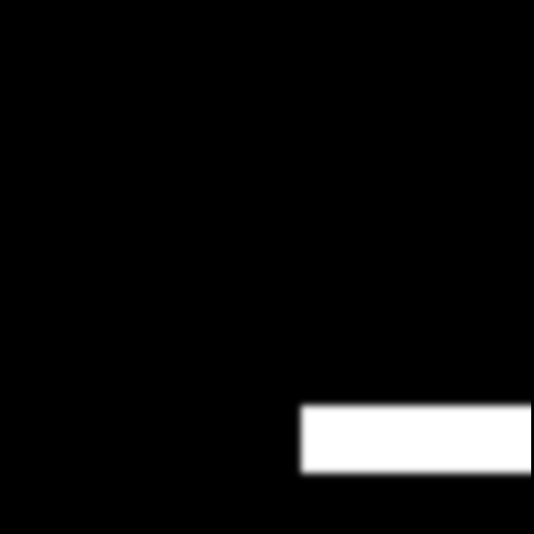}
    \end{subfigure}
    \begin{subfigure}{\fivecolsfactor\columnwidth}
        \includegraphics[width=\linewidth]{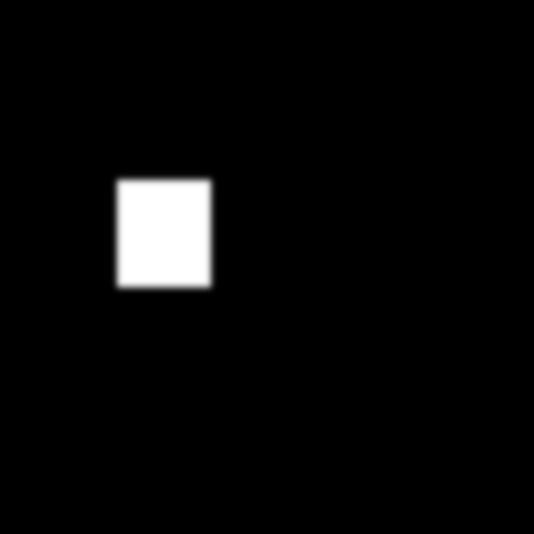}
    \end{subfigure}
    \begin{subfigure}{\fivecolsfactor\columnwidth}
        \includegraphics[width=\linewidth]{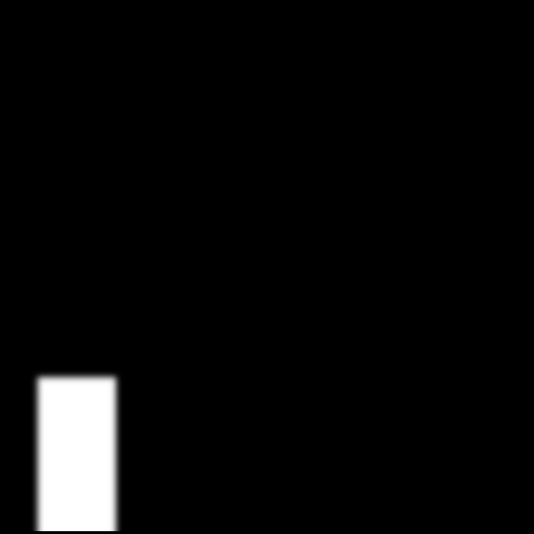}
    \end{subfigure}
    \begin{subfigure}{\fivecolsfactor\columnwidth}
        \includegraphics[width=\linewidth]{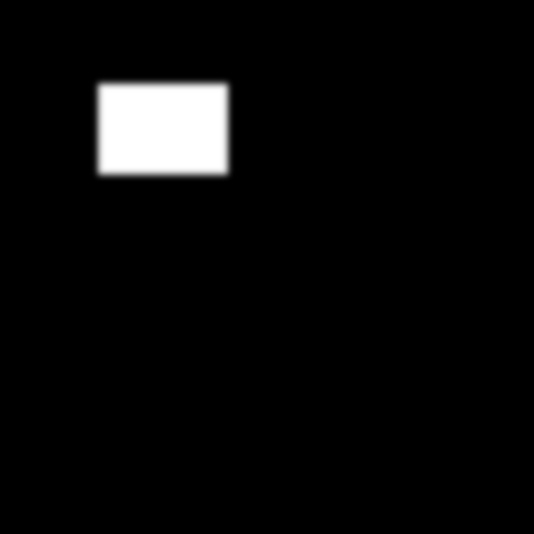}
    \end{subfigure}
    \begin{subfigure}{\fivecolsfactor\columnwidth}
        \includegraphics[width=\linewidth]{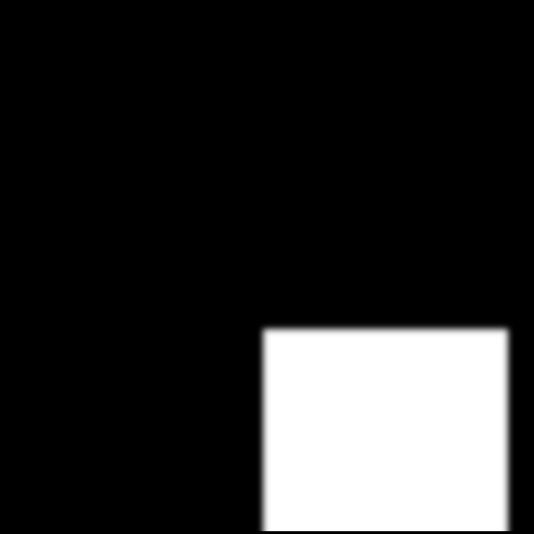}
    \end{subfigure}
    \begin{subfigure}{\fivecolsfactor\columnwidth}
        \includegraphics[width=\linewidth]{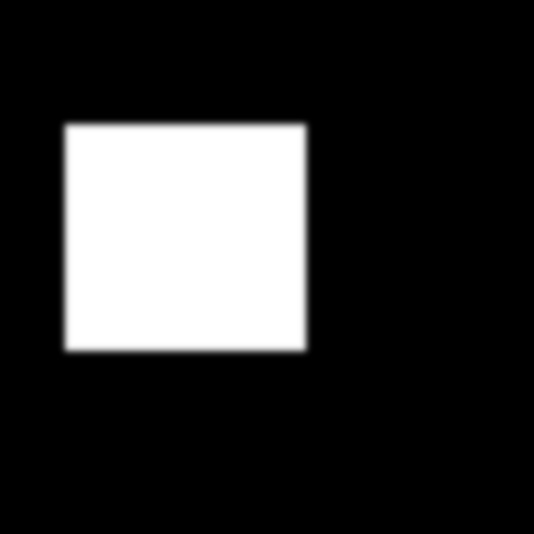}
    \end{subfigure}
    \begin{subfigure}{\fivecolsfactor\columnwidth}
        \includegraphics[width=\linewidth]{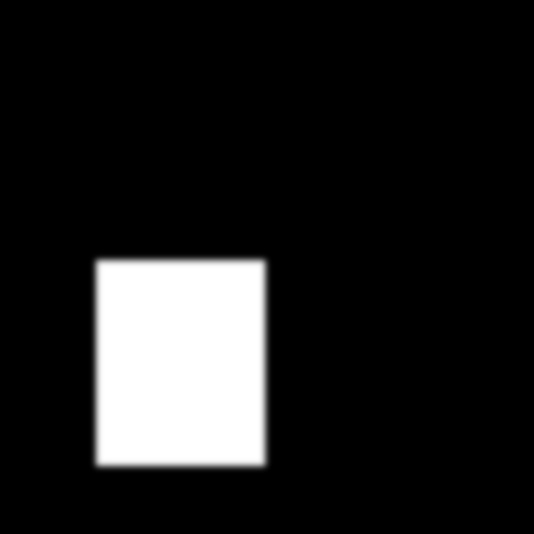}
    \end{subfigure}
    \begin{subfigure}{\fivecolsfactor\columnwidth}
        \includegraphics[width=\linewidth]{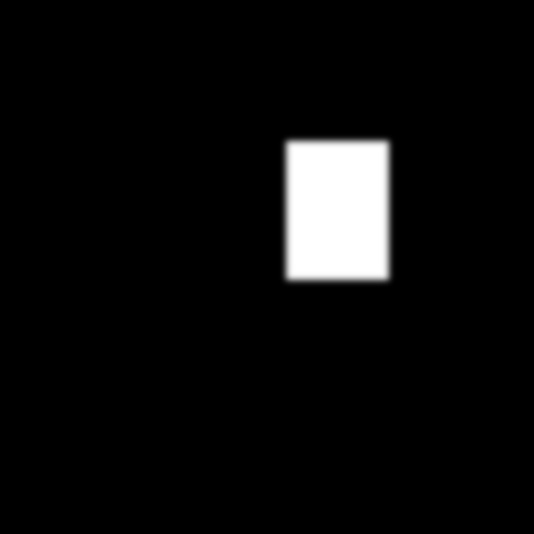}
    \end{subfigure}
    \begin{subfigure}{\fivecolsfactor\columnwidth}
        \includegraphics[width=\linewidth]{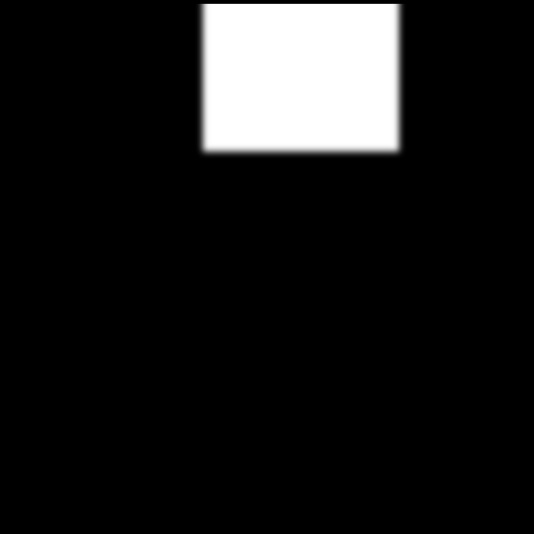}
    \end{subfigure}
    \hfill

    \begin{subfigure}{\fivecolsfactor\columnwidth}
        \includegraphics[width=\linewidth]{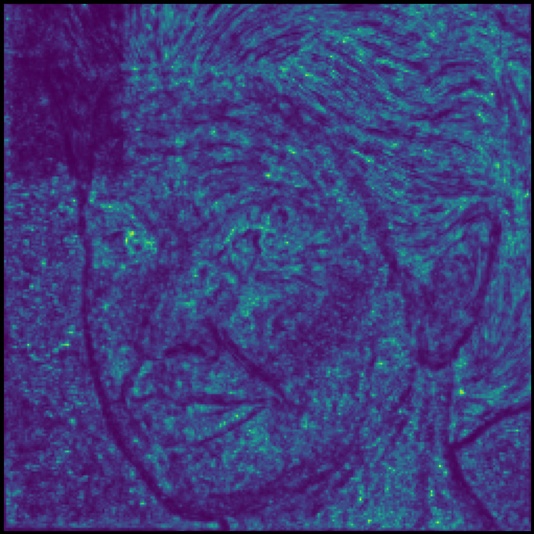}
    \end{subfigure}
    \begin{subfigure}{\fivecolsfactor\columnwidth}
        \includegraphics[width=\linewidth]{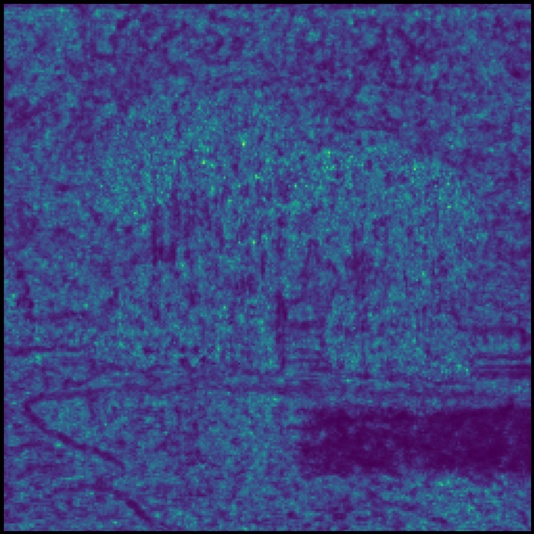}
    \end{subfigure}
    \begin{subfigure}{\fivecolsfactor\columnwidth}
        \includegraphics[width=\linewidth]{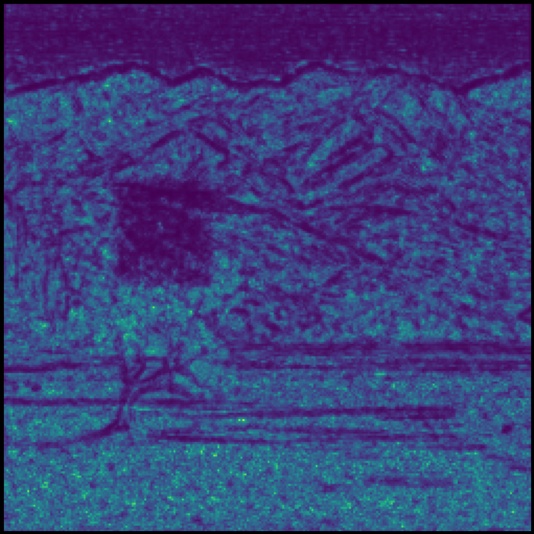}
    \end{subfigure}
    \begin{subfigure}{\fivecolsfactor\columnwidth}
        \includegraphics[width=\linewidth]{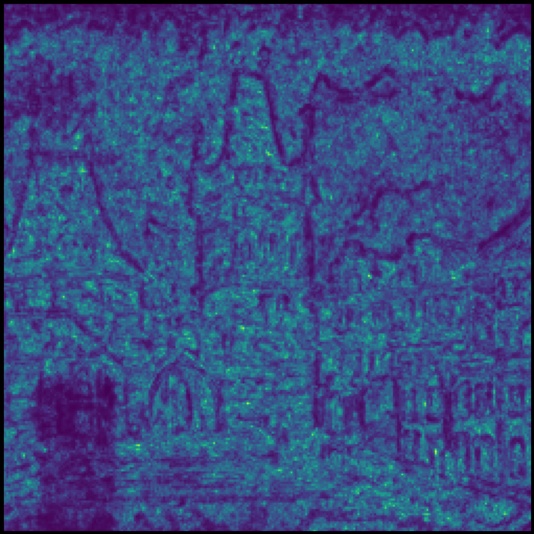}
    \end{subfigure}
    \begin{subfigure}{\fivecolsfactor\columnwidth}
        \includegraphics[width=\linewidth]{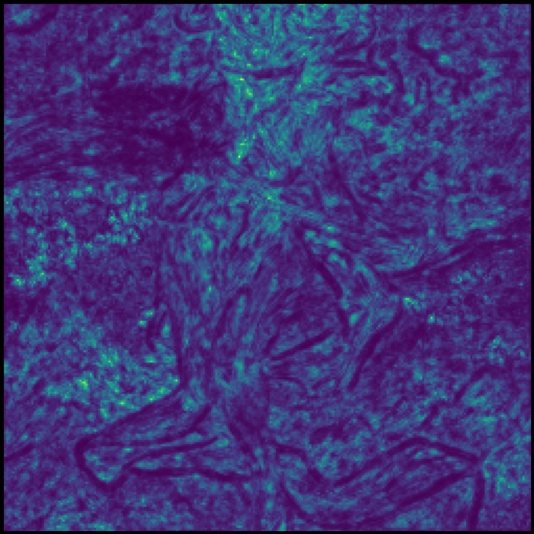}
    \end{subfigure}
        \begin{subfigure}{\fivecolsfactor\columnwidth}
        \includegraphics[width=\linewidth]{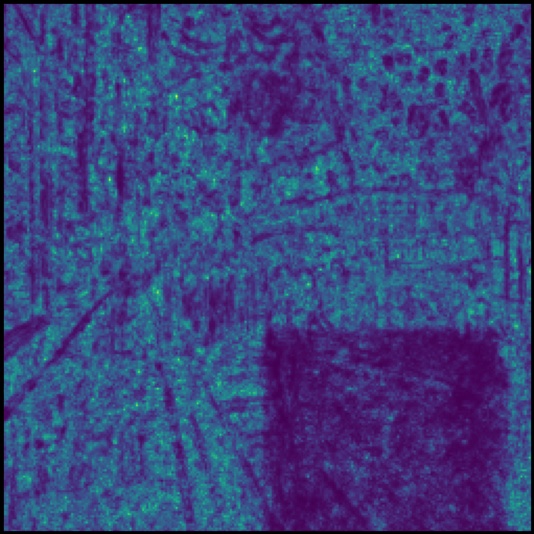}
    \end{subfigure}
    \begin{subfigure}{\fivecolsfactor\columnwidth}
        \includegraphics[width=\linewidth]{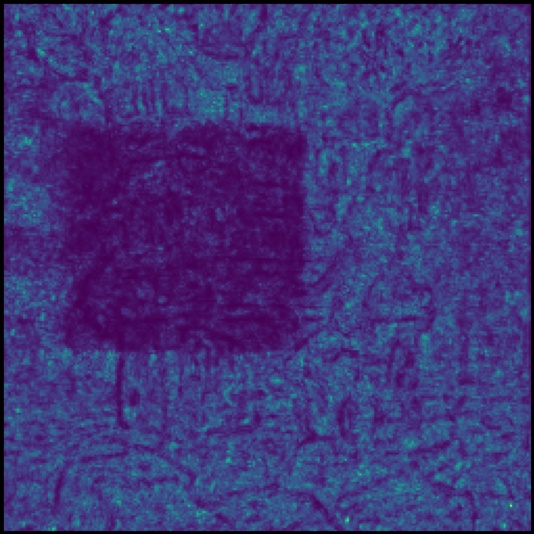}
    \end{subfigure}
    \begin{subfigure}{\fivecolsfactor\columnwidth}
        \includegraphics[width=\linewidth]{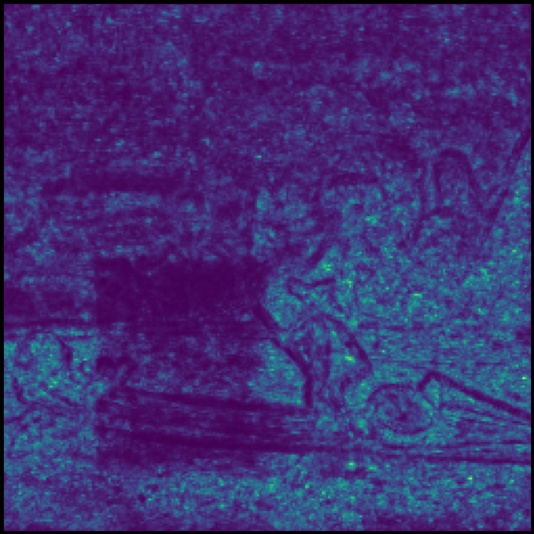}
    \end{subfigure}
    \begin{subfigure}{\fivecolsfactor\columnwidth}
        \includegraphics[width=\linewidth]{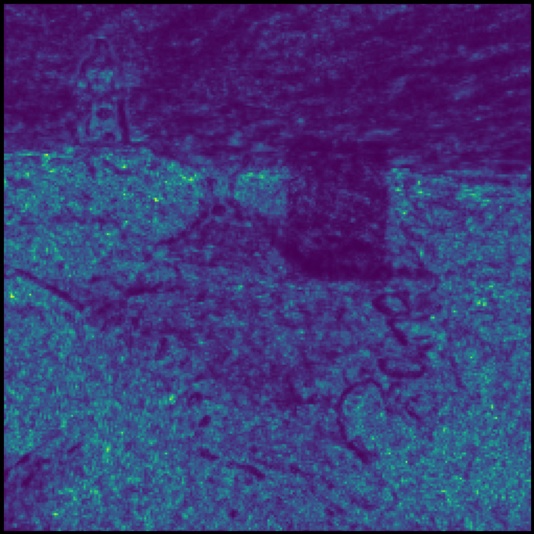}
    \end{subfigure}
    \begin{subfigure}{\fivecolsfactor\columnwidth}
        \includegraphics[width=\linewidth]{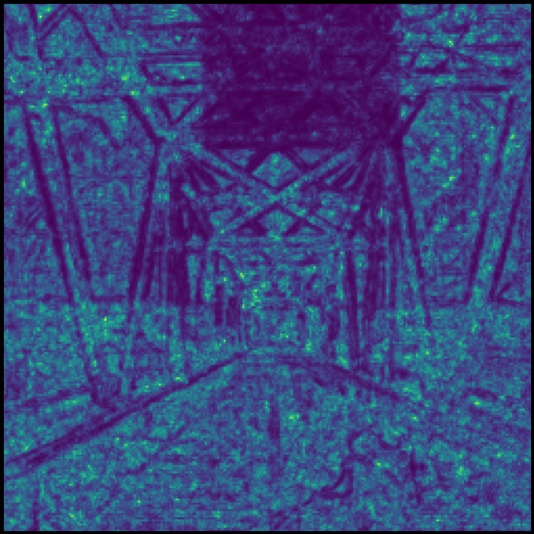}
    \end{subfigure}
    \hfill
    \caption{Additional examples illustrating the localization of inpainted regions using the reconstruction error. We show the images inpainted with Stable Diffusion 1.5 (top), the masks used for inpainting (center) and the reconstruction error maps (bottom), computed with $\text{LPIPS}_2$ and the AE from Stable Diffusion 1.5.}
    \label{fig:inpainting_examples_extended}
\end{figure}

\clearpage
\subsection{Robustness to Perturbations}\label{sup:robustness}
\begin{figure}[h]
    \centering
    \begin{subfigure}{0.18\textwidth}
        \includegraphics[width=\linewidth]{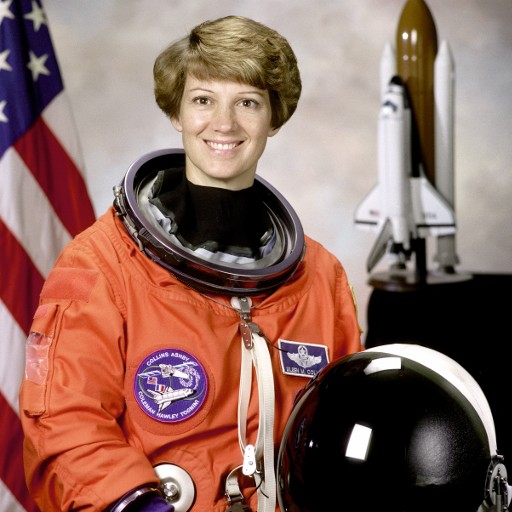}
        \caption{JPEG ($q=90$)}
    \end{subfigure}
    \begin{subfigure}{0.18\textwidth}
        \includegraphics[width=\linewidth]{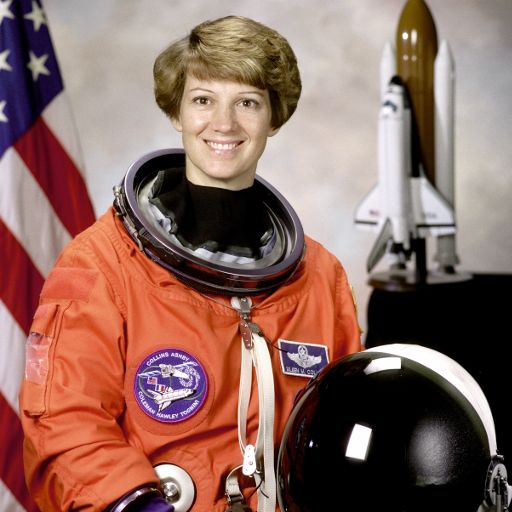}
        \caption{JPEG ($q=80$)}
    \end{subfigure}
    \begin{subfigure}{0.18\textwidth}
        \includegraphics[width=\linewidth]{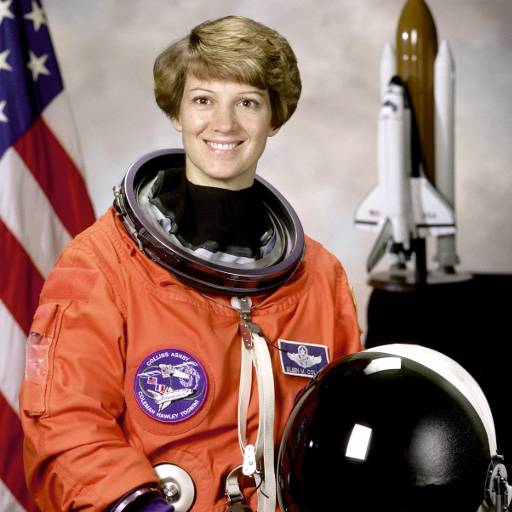}
        \caption{JPEG ($q=70$)}
    \end{subfigure}
    \begin{subfigure}{0.18\textwidth}
        \includegraphics[width=\linewidth]{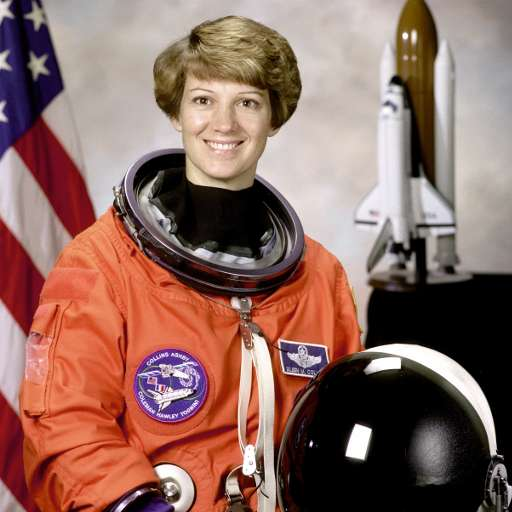}
        \caption{JPEG ($q=60$)}
    \end{subfigure}
    \begin{subfigure}{0.18\textwidth}
        \includegraphics[width=\linewidth]{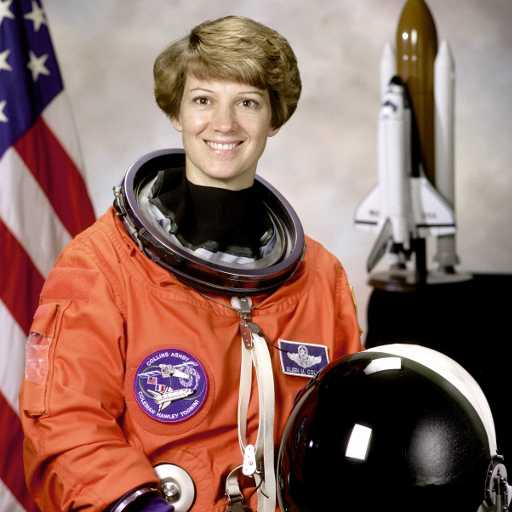}
        \caption{JPEG ($q=50$)}
    \end{subfigure}
    \hfill
    \begin{subfigure}{0.18\textwidth}
        \includegraphics[width=\linewidth]{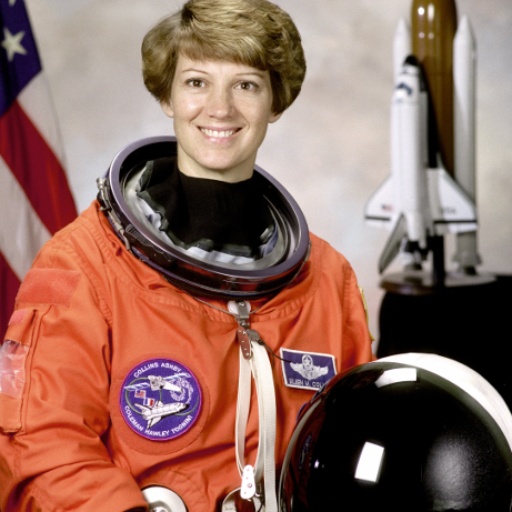}
        \caption{Crop ($f=0.9$)}
    \end{subfigure}
    \begin{subfigure}{0.18\textwidth}
        \includegraphics[width=\linewidth]{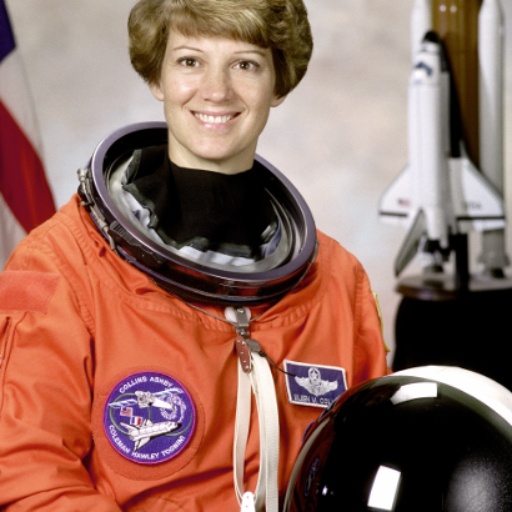}
        \caption{Crop ($f=0.8$)}
    \end{subfigure}
    \begin{subfigure}{0.18\textwidth}
        \includegraphics[width=\linewidth]{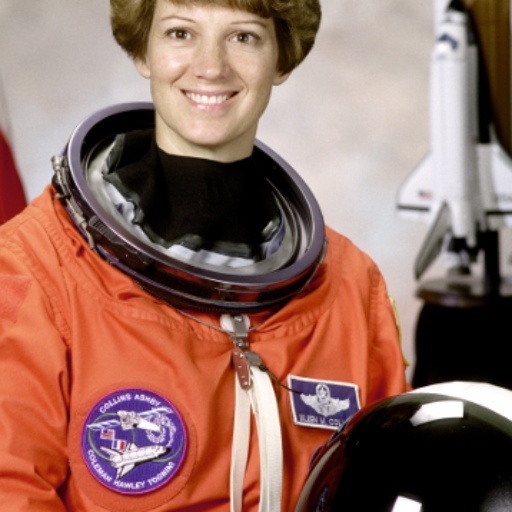}
        \caption{Crop ($f=0.7$)}
    \end{subfigure}
    \begin{subfigure}{0.18\textwidth}
        \includegraphics[width=\linewidth]{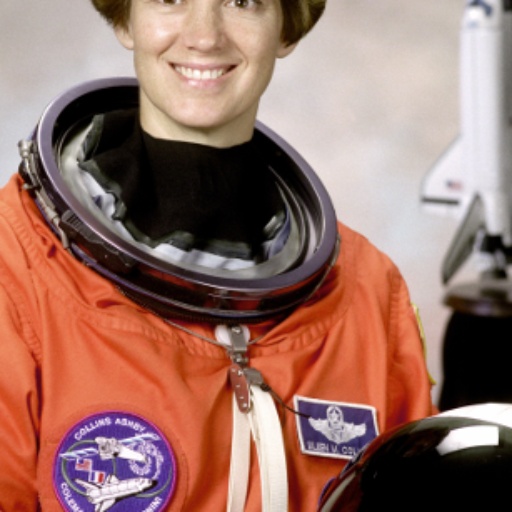}
        \caption{Crop ($f=0.6$)}
    \end{subfigure}
        \begin{subfigure}{0.18\textwidth}
        \includegraphics[width=\linewidth]{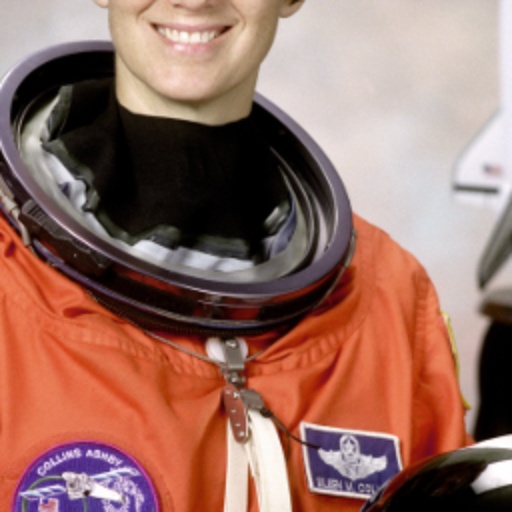}
        \caption{Crop ($f=0.5$)}
    \end{subfigure}
    \hfill
    \begin{subfigure}{0.18\textwidth}
        \includegraphics[width=\linewidth]{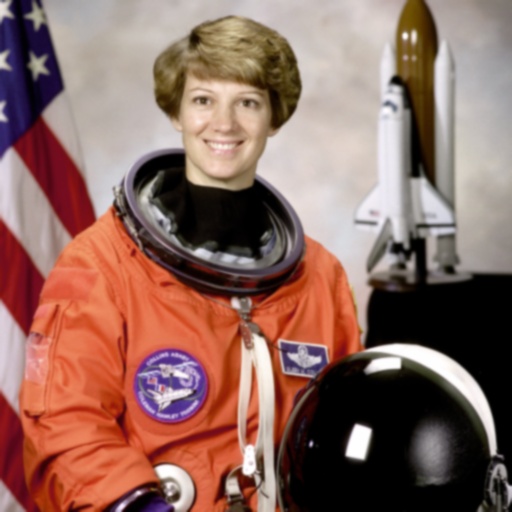}
        \caption{Blur ($\sigma=1.0$)}
    \end{subfigure}
    \begin{subfigure}{0.18\textwidth}
        \includegraphics[width=\linewidth]{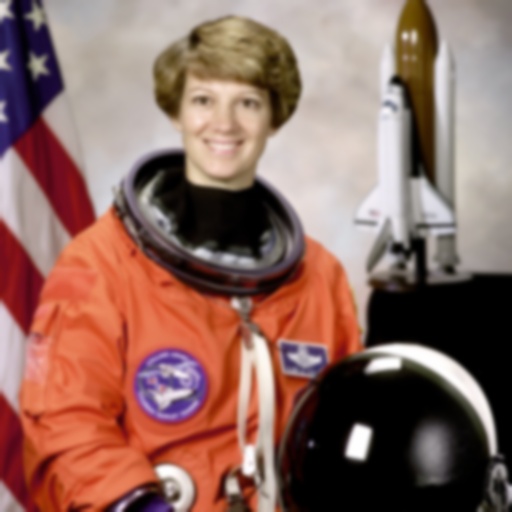}
        \caption{Blur ($\sigma=2.0$)}
    \end{subfigure}
    \begin{subfigure}{0.18\textwidth}
        \includegraphics[width=\linewidth]{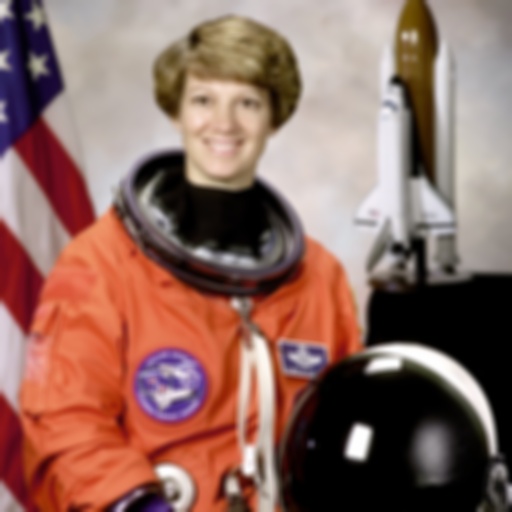}
        \caption{Blur ($\sigma=3.0$)}
    \end{subfigure}
    \begin{subfigure}{0.18\textwidth}
        \includegraphics[width=\linewidth]{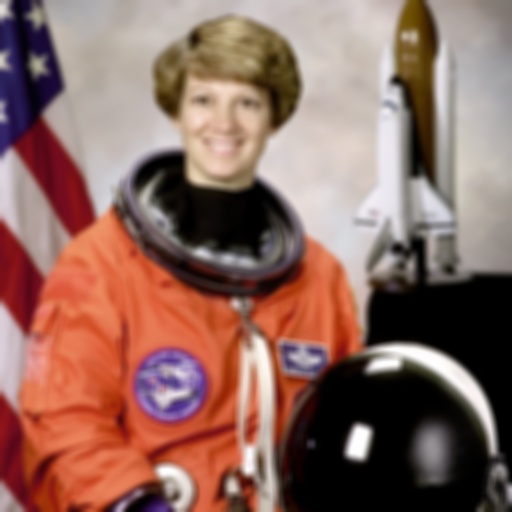}
        \caption{Blur ($\sigma=4.0$)}
    \end{subfigure}
    \begin{subfigure}{0.18\textwidth}
        \includegraphics[width=\linewidth]{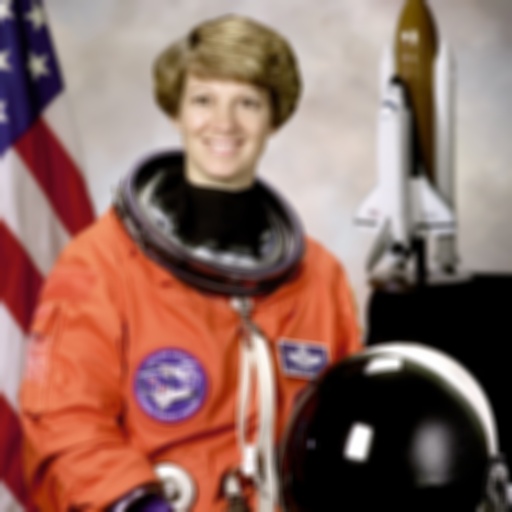}
        \caption{Blur ($\sigma=5.0$)}
    \end{subfigure}
    \hfill
    \begin{subfigure}{0.18\textwidth}
        \includegraphics[width=\linewidth]{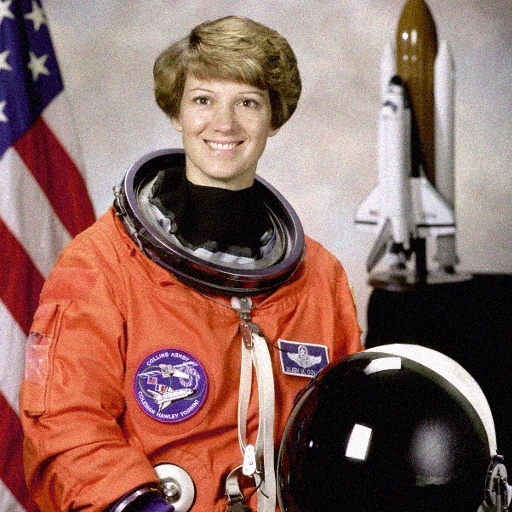}
        \caption{Noise ($\sigma=0.05$)}
    \end{subfigure}
    \begin{subfigure}{0.18\textwidth}
        \includegraphics[width=\linewidth]{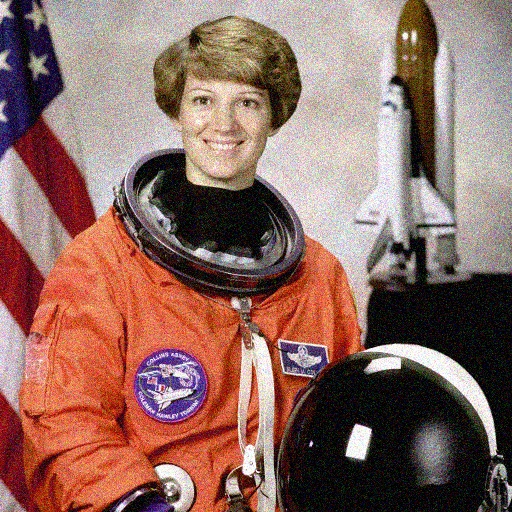}
        \caption{Noise ($\sigma=0.1$)}
    \end{subfigure}
    \begin{subfigure}{0.18\textwidth}
        \includegraphics[width=\linewidth]{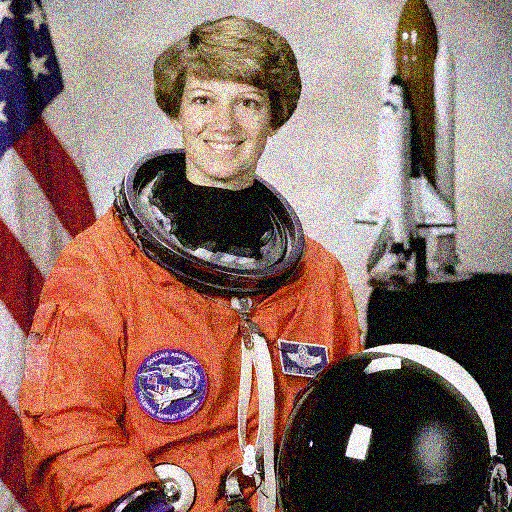}
        \caption{Noise ($\sigma=0.15$)}
    \end{subfigure}
    \begin{subfigure}{0.18\textwidth}
        \includegraphics[width=\linewidth]{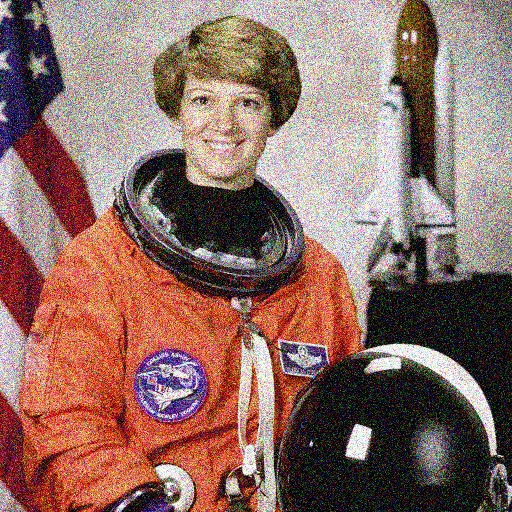}
        \caption{Noise ($\sigma=0.2$)}
    \end{subfigure}
    \begin{subfigure}{0.18\textwidth}
        \includegraphics[width=\linewidth]{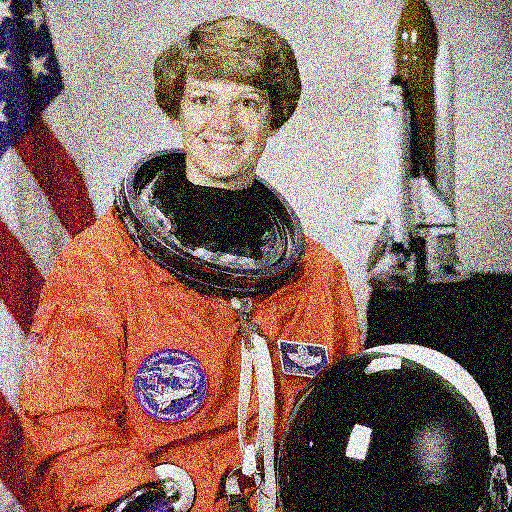}
        \caption{Noise ($\sigma=0.25$)}
    \end{subfigure}
    \caption{Visualization of different perturbations. In (a)-(e), $q$ denotes the JPEG quality factor. In (f)-(j) $f$ denotes the crop factor. In (k)-(t), $\sigma$ denotes the standard deviation of the Gaussian blur and noise, respectively.}
    \label{fig:perturbation_examples}
\end{figure}

In \cref{fig:perturbation_examples} we first provide examples that illustrate how the perturbed images evaluated in \cref{sec:experiments:additional} look like. Note that for JPEG and cropping, a smaller value ($q$ or $f$) leads to a stronger perturbation, while for the addition of noise and blurring a larger value ($\sigma$) has a stronger effect. For blurring we use a kernel size of $9$.

\begin{figure}
    \centering
    \begin{subfigure}{0.245\linewidth}
        \centering
        \includegraphics{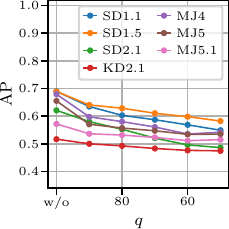}
        \caption{JPEG}
    \end{subfigure}
    \begin{subfigure}{0.245\linewidth}
        \centering
        \includegraphics{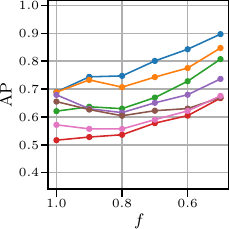}
        \caption{Crop}
    \end{subfigure}
    \begin{subfigure}{0.245\linewidth}
        \centering
        \includegraphics{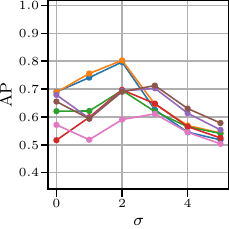}
        \caption{Blur}
    \end{subfigure}
    \begin{subfigure}{0.245\linewidth}
        \centering
        \includegraphics{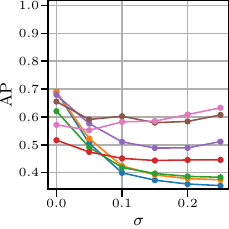}
        \caption{Noise}
    \end{subfigure}
    \caption{Detection performance of the detector from \citet{gragnanielloAreGANGenerated2021} on perturbed images, measured in AP.}
    \label{fig:perturbations_grag}
\end{figure}

\begin{figure}
    \centering
    \begin{subfigure}{0.245\linewidth}
        \centering
        \includegraphics{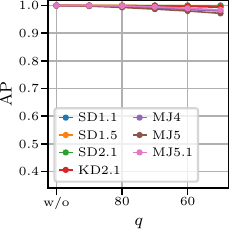}
        \caption{JPEG}
    \end{subfigure}
    \begin{subfigure}{0.245\linewidth}
        \centering
        \includegraphics{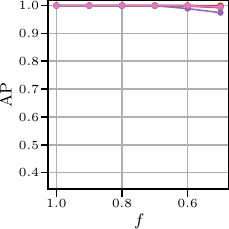}
        \caption{Crop}
    \end{subfigure}
    \begin{subfigure}{0.245\linewidth}
        \centering
        \includegraphics{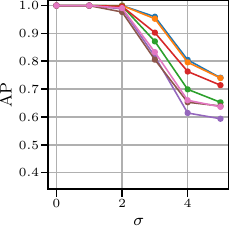}
        \caption{Blur}
    \end{subfigure}
    \begin{subfigure}{0.245\linewidth}
        \centering
        \includegraphics{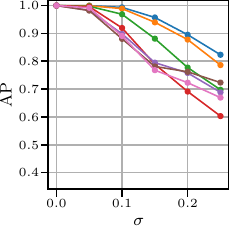}
        \caption{Noise}
    \end{subfigure}
    \caption{Detection performance of the detector from \citet{corviDetectionSyntheticImages2023} on perturbed images, measured in AP.}
    \label{fig:perturbations_corvi}
\end{figure}

\begin{figure}
    \centering
    \begin{subfigure}{0.245\linewidth}
        \centering
        \includegraphics{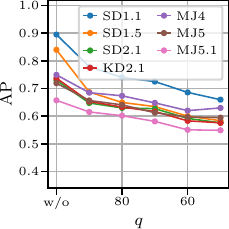}
        \caption{JPEG}
    \end{subfigure}
    \begin{subfigure}{0.245\linewidth}
        \centering
        \includegraphics{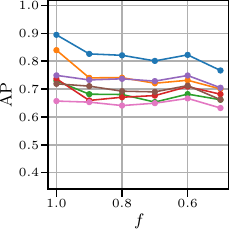}
        \caption{Crop}
    \end{subfigure}
    \begin{subfigure}{0.245\linewidth}
        \centering
        \includegraphics{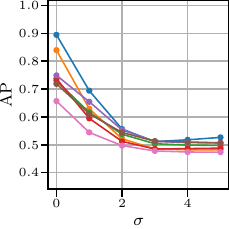}
        \caption{Blur}
    \end{subfigure}
    \begin{subfigure}{0.245\linewidth}
        \centering
        \includegraphics{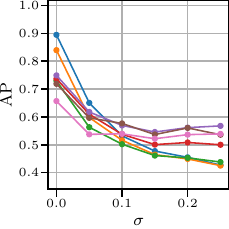}
        \caption{Noise}
    \end{subfigure}
    \caption{Detection performance of the detector from \citet{ojhaUniversalFakeImage2023} on perturbed images, measured in AP.}
    \label{fig:perturbations_ojha}
\end{figure}

\begin{figure}
    \centering
    \begin{subfigure}{0.245\linewidth}
        \centering
        \includegraphics{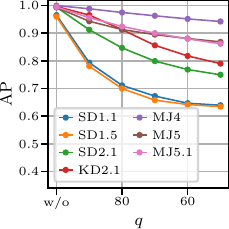}
        \caption{JPEG}
    \end{subfigure}
    \begin{subfigure}{0.245\linewidth}
        \centering
        \includegraphics{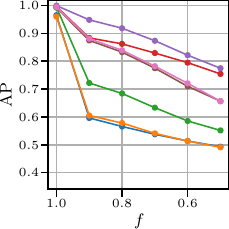}
        \caption{Crop}
    \end{subfigure}
    \begin{subfigure}{0.245\linewidth}
        \centering
        \includegraphics{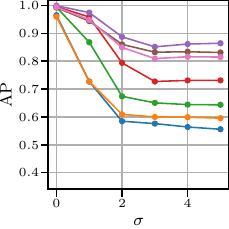}
        \caption{Blur}
    \end{subfigure}
    \begin{subfigure}{0.245\linewidth}
        \centering
        \includegraphics{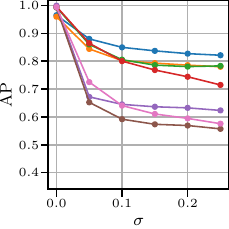}
        \caption{Noise}
    \end{subfigure}
    \caption{Detection performance of \method{} (with LPIPS and $\Delta_\text{Min}$) on perturbed images, measured in AP.}
    \label{fig:perturbations_sum}
\end{figure}

\begin{figure}
    \centering
    \begin{subfigure}{0.245\linewidth}
        \centering
        \includegraphics{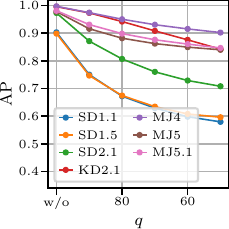}
        \caption{JPEG}
    \end{subfigure}
    \begin{subfigure}{0.245\linewidth}
        \centering
        \includegraphics{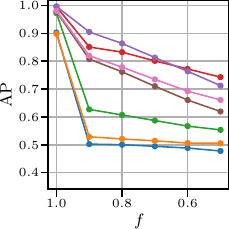}
        \caption{Crop}
    \end{subfigure}
    \begin{subfigure}{0.245\linewidth}
        \centering
        \includegraphics{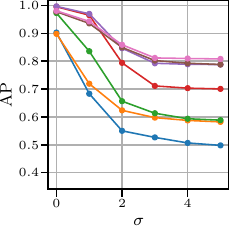}
        \caption{Blur}
    \end{subfigure}
    \begin{subfigure}{0.245\linewidth}
        \centering
        \includegraphics{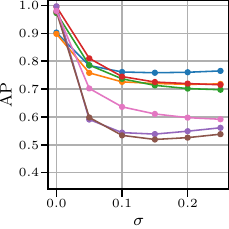}
        \caption{Noise}
    \end{subfigure}
    \caption{Detection performance of \method{} (with $\text{LPIPS}_1$ and $\Delta_\text{Min}$) on perturbed images, measured in AP.}
    \label{fig:perturbations_conv1}
\end{figure}

\begin{figure}
    \centering
    \begin{subfigure}{0.245\linewidth}
        \centering
        \includegraphics{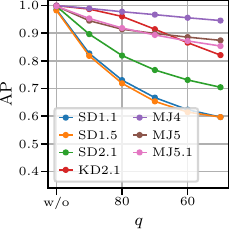}
        \caption{JPEG}
    \end{subfigure}
    \begin{subfigure}{0.245\linewidth}
        \centering
        \includegraphics{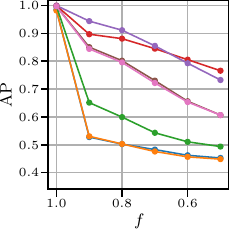}
        \caption{Crop}
    \end{subfigure}
    \begin{subfigure}{0.245\linewidth}
        \centering
        \includegraphics{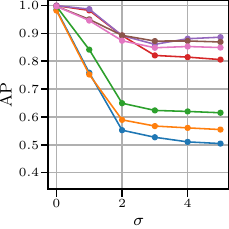}
        \caption{Blur}
    \end{subfigure}
    \begin{subfigure}{0.245\linewidth}
        \centering
        \includegraphics{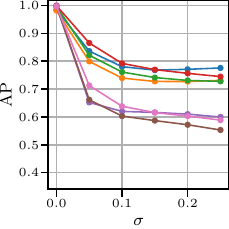}
        \caption{Noise}
    \end{subfigure}
    \caption{Detection performance of \method{} (with $\text{LPIPS}_2$ and $\Delta_\text{Min}$) on perturbed images, measured in AP.}
    \label{fig:perturbations_conv2}
\end{figure}

\begin{figure}
    \centering
    \begin{subfigure}{0.245\linewidth}
        \centering
        \includegraphics{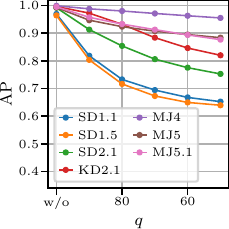}
        \caption{JPEG}
    \end{subfigure}
    \begin{subfigure}{0.245\linewidth}
        \centering
        \includegraphics{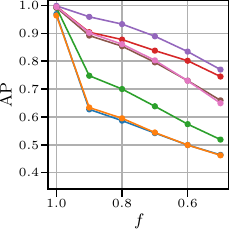}
        \caption{Crop}
    \end{subfigure}
    \begin{subfigure}{0.245\linewidth}
        \centering
        \includegraphics{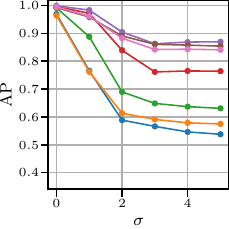}
        \caption{Blur}
    \end{subfigure}
    \begin{subfigure}{0.245\linewidth}
        \centering
        \includegraphics{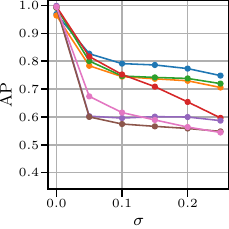}
        \caption{Noise}
    \end{subfigure}
    \caption{Detection performance of \method{} (with $\text{LPIPS}_3$ and $\Delta_\text{Min}$) on perturbed images, measured in AP.}
    \label{fig:perturbations_conv3}
\end{figure}

\begin{figure}
    \centering
    \begin{subfigure}{0.245\linewidth}
        \centering
        \includegraphics{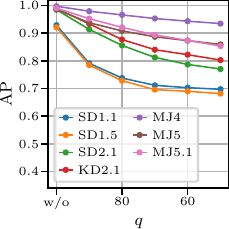}
        \caption{JPEG}
    \end{subfigure}
    \begin{subfigure}{0.245\linewidth}
        \centering
        \includegraphics{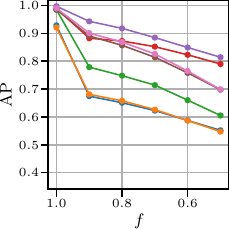}
        \caption{Crop}
    \end{subfigure}
    \begin{subfigure}{0.245\linewidth}
        \centering
        \includegraphics{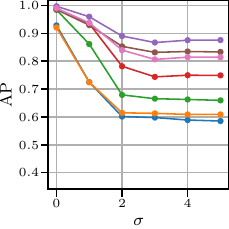}
        \caption{Blur}
    \end{subfigure}
    \begin{subfigure}{0.245\linewidth}
        \centering
        \includegraphics{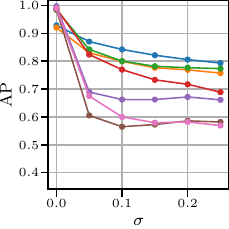}
        \caption{Noise}
    \end{subfigure}
    \caption{Detection performance of \method{} (with $\text{LPIPS}_4$ and $\Delta_\text{Min}$) on perturbed images, measured in AP.}
    \label{fig:perturbations_conv4}
\end{figure}

\begin{figure}
    \centering
    \begin{subfigure}{0.245\linewidth}
        \centering
        \includegraphics{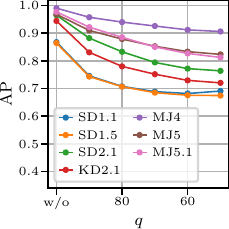}
        \caption{JPEG}
    \end{subfigure}
    \begin{subfigure}{0.245\linewidth}
        \centering
        \includegraphics{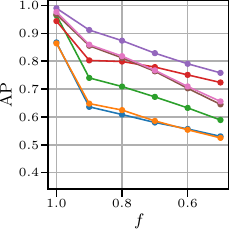}
        \caption{Crop}
    \end{subfigure}
    \begin{subfigure}{0.245\linewidth}
        \centering
        \includegraphics{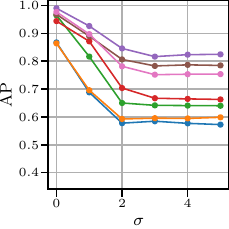}
        \caption{Blur}
    \end{subfigure}
    \begin{subfigure}{0.245\linewidth}
        \centering
        \includegraphics{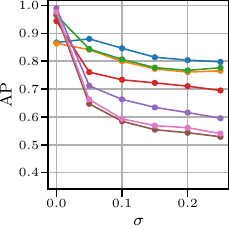}
        \caption{Noise}
    \end{subfigure}
    \caption{Detection performance of \method{} (with $\text{LPIPS}_5$ and $\Delta_\text{Min}$) on perturbed images, measured in AP.}
    \label{fig:perturbations_conv5}
\end{figure}

\begin{figure}
    \centering
    \begin{subfigure}{0.245\linewidth}
        \centering
        \includegraphics{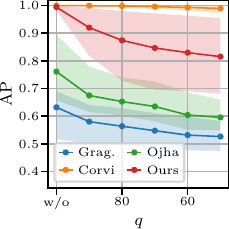}
        \caption{JPEG}
    \end{subfigure}
    \begin{subfigure}{0.245\linewidth}
        \centering
        \includegraphics{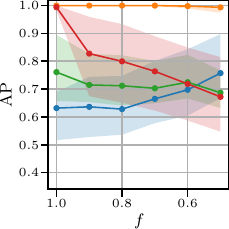}
        \caption{Crop}
    \end{subfigure}
    \begin{subfigure}{0.245\linewidth}
        \centering
        \includegraphics{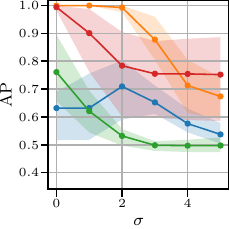}
        \caption{Blur}
    \end{subfigure}
    \begin{subfigure}{0.245\linewidth}
        \centering
        \includegraphics{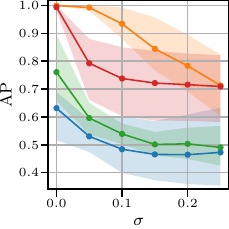}
        \caption{Noise}
    \end{subfigure}
    \caption{Detection performance of \method{} (with optimal LPIPS variant for each setting and $\Delta_\text{Min}$) and baselines on perturbed images, measured in AP. Results are averaged over all datasets, with shaded areas indicating the minimum and maximum.}
    \label{fig:perturbations_best_layer}
\end{figure}

In addition to \cref{fig:perturbations_ap}, we provide the results on individual datasets for the baselines (see \cref{fig:perturbations_grag,fig:perturbations_corvi,fig:perturbations_ojha}) as well as all variants of LPIPS (see \cref{fig:perturbations_sum,fig:perturbations_conv1,fig:perturbations_conv2,fig:perturbations_conv3,fig:perturbations_conv4,fig:perturbations_conv5}). We observe that in some settings, higher layers perform better than $\text{LPIPS}_2$. We suppose that higher layers are less affected by the loss of details caused by the perturbations. This is illustrated by an alternative version of \cref{fig:perturbations_ap} where we select the optimal LPIPS layer for each perturbation (see \cref{fig:perturbations_best_layer}). While this of course gives an unfair advantage, it shows the potential robustness of \method{}.

\clearpage

\clearpage
\section{Analysis of DIRE}
\label{sec:dire_analysis}
While trying to reproduce the results from \citet{wangDIREDiffusiongeneratedImage2023}, we found that the experiments conducted by the authors might have been affected by an unwanted bias in the data, as we briefly mention in \cref{sec:experiments:comparison}.
In this section, we provide details on the experiments and results that led us to our findings.

\subsection{DIRE}
Before proceeding, we briefly summarize the approach proposed by \citet{wangDIREDiffusiongeneratedImage2023}.
The idea behind DIRE is that a diffusion model should be able to reconstruct an image more accurately (compared to the original image), if that image is generated by a diffusion model, as opposed to being a real image.
In their analysis, the authors use a diffusion model (ADM~\citep{dhariwalDiffusionModelsBeat2021} in most of their experiments) to invert a given image $x_0$ to its initial time step $x_T$ (corresponding to Gaussian noise) using the inverted DDIM sampler~\citep{songDenoisingDiffusionImplicit2022}, and then denoise back to $x'_0$ using DDIM as usual.
The authors hypothesize that the pixel-wise difference between $x_0$ and $x'_0$, the DIRE representation, has different characteristics for real and generated images.
To exploit this for classification, they train a binary classifier on the pixel-wise reconstruction error images $|x_0 - x'_0|$ (from real images and images generated using some diffusion model).
The authors provide code, data and pre-trained models\footnote{\url{https://github.com/ZhendongWang6/DIRE}},
which we use according to the official instructions in the following experiments.

In the following tables, we test different pre-trained classifiers, which all use the ResNet-50 architecture but are trained on different datasets.
These classifiers are ADM-B (trained on LSUN Bedroom vs images generated by ADM trained on LSUN Bedroom), ADM-IN (trained on ImageNet vs images generated by ADM trained on ImageNet), PNDM-B, and IDDPM-B.
Further, when referring to datasets (such as in the second and third column in the tables), LSUN-B stands for LSUN Bedroom and ADM-B and ADM-IN stand for images generated using ADM trained on LSUN Bedroom and ImageNet, respectively.

\subsection{Experiment 1}
We first experiment with the LSUN Bedroom test set provided by the authors.
We obtain the images by downloading and extracting \texttt{\small adm.tar.gz} (images generated by ADM trained on LSUN Bedroom) and \texttt{\small real.tar.gz} (real LSUN Bedroom images) from \texttt{\small dire/test/lsun\_bedroom/lsun\_bedroom}.
It should be noted that the authors provide original images, reconstructions, and the DIRE representations (i.e., the differences between originals and reconstructions). The script \texttt{\small test.py}, which is used to evaluate the classification performance, directly uses the DIRE representations.

The key observation we make is that the DIREs of fake images are stored as PNG files (which uses lossless compression), while the DIREs of real images are stored as JPEG files (compressed with quality factor $95$).
We suspect the reason for this is in \texttt{\small guided\_diffusion/compute\_dire.py}, where the DIRE files inherit the extension of the corresponding original input image file and the image library automatically applies the corresponding compression.
Since the original real images were provided in JPEG, their DIRE images are also automatically stored with JPEG compression.
We stress that the authors could not have avoided the compression of the original real images, since the dataset simply comes in this format. However, saving the DIRE in the lossy JPEG format introduces an unwanted bias, as our experiments show.
A possible solution to this would have been to store the generated images using the same JPEG compression as the original real images (since uncompressed real images are not available) before computing their DIREs.

\paragraph{Variant 1} We first run \texttt{\small test.py} according to the authors' instructions given in the repository. We compare the performance of different pre-trained classifiers and report the results in \cref{tab:adm_orig}. The ``Classifier'' column denotes on which kind of generated images (together with the corresponding real images) the classifier was trained on. In this setting, we are able to reproduce the authors' results: all classifiers achieve very good detection and even generalize to other datasets.

\paragraph{Variant 2}
We repeat the previous experiment, but this time we ensure that the DIREs of real and generated images are saved consistently. Since we do not have access to real DIREs in their uncompressed form, we instead convert the fake DIREs to JPEG with the same quality level ($95$) as the real DIREs provided by the authors.

As the results in \cref{tab:adm_jpeg95} show, the performance of all classifiers drops from almost perfect detection (see \cref{tab:adm_orig}) to random guessing. In particular, all images are classified as being real. We emphasize that the only change we made (compared to the previous evaluation) was to convert the uncompressed DIREs of generated images to the same format as the DIREs of real images.

\begin{table}[tbh]
    \setlength{\tabcolsep}{4.0pt}
    \begin{subtable}{0.49\columnwidth}
        \centering
        \scriptsize
        \begin{tabular}{@{\ }l l l | r r r r@{\ }}
            \toprule
            Classifier & Real (Test) & Fake (Test) & Acc & AP & $\text{Acc}_R$ & $\text{Acc}_F$ \\
            \midrule
            ADM-B & LSUN-B & ADM-B & 1.000 & 1.000 & 1.000 & 1.000 \\
            PNDM-B & LSUN-B & ADM-B & 1.000 & 1.000 & 1.000 & 1.000 \\
            IDDPM-B & LSUN-B & ADM-B & 0.996 & 1.000 & 1.000 & 0.991 \\
            ADM-IN & ImageNet & ADM-IN & 0.999 & 1.000 & 1.000 & 0.998 \\
            ADM-B & ImageNet & ADM-IN & 0.998 & 1.000 & 0.999 & 0.999 \\
            \bottomrule
        \end{tabular}
        \caption{real DIREs: JPEG, fake DIREs: PNG
        }
        \label{tab:adm_orig}
    \end{subtable}
    \begin{subtable}{0.49\columnwidth}
        \centering
        \scriptsize
        \begin{tabular}{@{\ }l l l | r r r r@{\ }}
            \toprule
            Classifier & Real (Test) & Fake (Test) & Acc & AP & $\text{Acc}_R$ & $\text{Acc}_F$ \\
            \midrule
            ADM-B & LSUN-B & ADM-B & 0.501 & 0.660 & 1.000 & 0.200 \\
            PNDM-B & LSUN-B & ADM-B & 0.500 & 0.660 & 1.000 & 0.000 \\
            IDDPM-B & LSUN-B & ADM-B & 0.502 & 0.670 & 1.000 & 0.400 \\
            ADM-IN & ImageNet & ADM-IN & 0.500 & 0.670 & 1.000 & 0.100 \\
            ADM-B & ImageNet & ADM-IN & 0.500 & 0.520 & 0.999 & 0.000 \\
            \bottomrule
        \end{tabular}
        \caption{real DIREs: JPEG, fake DIREs: JPEG
        }
        \label{tab:adm_jpeg95}
    \end{subtable}
    \caption{Performance of different pre-trained classifiers on test data provided by the authors. The only difference between (a) and (b) is that in (b) the reconstructions of fake images are stored in the JPEG format (quality level $95$), like the real images.}
\end{table}

\subsection{Experiment 2}
In a second experiment we further demonstrate that the format in which the DIREs are saved in has a significant influence on the detection. We take \num{1000} DIREs of images generated by ADM (trained on LSUN Bedroom) (\texttt{dire/test/lsun\_bedroom/lsun\_bedroom/adm.tar.gz}) and split them in two sets A and B, each containing $500$ images.

\paragraph{Variant 1}
All DIREs are stored in the lossless PNG format. We pretend that set A is actually a collection of real images, while set B is considered to be fake. Again, we use \texttt{\small test.py} provided by the authors to evaluate the detection performance. The results in \cref{tab:adm_allfake_orig} are as expected, since all images are actually generated and half of them have the wrong label, the classifiers achieve an accuracy of approximately $0.5$.

\paragraph{Variant 2}
We repeat the previous experiment, but convert all DIREs in set A (which we label as being real) to JPEG with quality $95$.
All other parameters remain unchanged. The results in \cref{tab:adm_allfake_jpeg95} show that set A (containing DIREs from generated images saved as JPEGs) is now classified as being real. Thus, the almost perfect classification accuracy (compared to \cref{tab:adm_allfake_orig}) is caused exclusively by the fact that we converted the DIREs in set A to the JPEG format.

\begin{table}[tbh]
    \setlength{\tabcolsep}{4.0pt}
    \begin{subtable}{0.49\columnwidth}
        \centering
        \scriptsize
        \begin{tabular}{@{\ }l l l | r r r r@{\ }}
            \toprule
            Classifier & Real* (Test) & Fake (Test) & Acc & AP & $\text{Acc}_R$ & $\text{Acc}_F$ \\
            \midrule
            ADM-B & ADM-B & ADM-B & 0.500 & 0.500 & 0.000 & 1.000 \\
            PNDM-B & ADM-B & ADM-B & 0.500 & 0.490 & 0.000 & 1.000 \\
            IDDPM-B & ADM-B & ADM-B & 0.503 & 0.490 & 0.012 & 0.994 \\
            \bottomrule
        \end{tabular}
        \caption{real* DIREs: PNG, fake DIREs: PNG}
        \label{tab:adm_allfake_orig}
    \end{subtable}
    \begin{subtable}{0.49\columnwidth}
        \centering
        \scriptsize
        \begin{tabular}{@{\ }l l l | r r r r@{\ }}
            \toprule
            Classifier & Real* (Test) & Fake (Test) & Acc & AP & $\text{Acc}_R$ & $\text{Acc}_F$ \\
            \midrule
            ADM-B & ADM-B & ADM-B & 0.999 & 1.000 & 0.998 & 1.000 \\
            PNDM-B & ADM-B & ADM-B & 0.997 & 1.000 & 0.994 & 1.000 \\
            IDDPM-B & ADM-B & ADM-B & 0.997 & 1.000 & 1.000 & 0.994 \\
            \bottomrule
        \end{tabular}
        \caption{real* DIREs: JPEG, fake DIREs: PNG}
        \label{tab:adm_allfake_jpeg95}
    \end{subtable}
    \caption{Performance of different pre-trained classifiers on test data provided by the authors. ``Real*'' indicates that the images are labeled as being real but are actually generated by ADM. The only difference between (a) and (b) is that in (b) the reconstructions of images that are labeled as being real are stored in the JPEG format (quality level $95$).}
\end{table}

\subsection{Discussion}
In both experiments we show that the classifiers provided by \citet{wangDIREDiffusiongeneratedImage2023} suffer from an unwanted bias in the training data. The DIREs of real images were saved as JPEGs, while those of generated images were saved as lossless PNGs. As a consequence, the detector is highly sensitive to the presence of compression artifacts. In other words, the format in which the DIREs are saved controls whether it is classified as being real or fake, not whether the image is actually real or fake. Based on our findings, we believe that the authors' experiments should be re-evaluated without the bias caused by the inconsistent file formats.

\end{document}